\documentclass[11pt]{article}
	
	\newcommand{\blind}{0}
	
    \makeatletter
    \renewcommand\section{\@startsection {section}{1}{\z@}%
                                       {-3.5ex \@plus -1ex \@minus -.2ex}%
                                       {2.3ex \@plus.2ex}%
                                       {\normalfont\fontfamily{phv}\fontsize{16}{19}\bfseries}}
    \renewcommand\subsection{\@startsection{subsection}{2}{\z@}%
                                         {-3.25ex\@plus -1ex \@minus -.2ex}%
                                         {1.5ex \@plus .2ex}%
                                         {\normalfont\fontfamily{phv}\fontsize{14}{17}\bfseries}}
    \renewcommand\subsubsection{\@startsection{subsubsection}{3}{\z@}%
                                        {-3.25ex\@plus -1ex \@minus -.2ex}%
                                         {1.5ex \@plus .2ex}%
                                         {\normalfont\normalsize\fontfamily{phv}\fontsize{14}{17}\selectfont}}
    \makeatother
	
	\usepackage{amsmath}
	\usepackage{graphicx}
	\usepackage{enumerate}
	\usepackage{xcolor}
	\usepackage{natbib} 
	\usepackage{url} 
	\newtheorem{definition}{Definition}   
        \newtheorem{proposition}{Proposition}

\usepackage{algorithm,algorithmic}
	
\begin{document}
		
		\def\spacingset#1{\renewcommand{\baselinestretch}%
			{#1}\small\normalsize} \spacingset{1}
		
		\if0\blind
		{
			\title{\bf Differentially Private Log-Location-Scale Regression Using Functional Mechanism}
			\author{Jiewen Sheng $^a$ and Xiaolei Fang $^b$ \\
			$^a$ Graduate Program in Operations Research, North Carolina State University\\
             $^b$Edward P. Fitts Department of Industrial and Systems Engineering, \\North Carolina State University }
			\date{}
			\maketitle
		} \fi
		
		\if1\blind
		{

            \title{\bf \emph{IISE Transactions} \LaTeX \ Template}
			\author{Author information is purposely removed for double-blind review}
			
\bigskip
			\bigskip
			\bigskip
			\begin{center}
				{\LARGE\bf \emph{IISE Transactions} \LaTeX \ Template}
			\end{center}
			\medskip
		} \fi
		\bigskip
		
	\begin{abstract}

This article introduces differentially private log-location-scale (DP-LLS) regression models, which incorporate differential privacy into LLS regression through the functional mechanism. The proposed models are established by injecting noise into the log-likelihood function of LLS regression for perturbed parameter estimation. We will derive the sensitivities utilized to determine the magnitude of the injected noise and prove that the proposed DP-LLS models satisfy $\epsilon$-differential privacy. In addition, we will conduct simulations and case studies to evaluate the performance of the proposed models. The findings suggest that predictor dimension, training sample size, and privacy budget are three key factors impacting the performance of the proposed DP-LLS regression models. Moreover, the results indicate that a sufficiently large training dataset is needed to simultaneously ensure decent performance of the proposed models and achieve a satisfactory level of privacy protection.

	\end{abstract}
			
	\noindent%

	\spacingset{1.5} 

\section{Introduction} \label{s:intro}
Log-location-scale (LLS) regression is a type of regression model that extends the standard linear regression model to accommodate situations where the response variable exhibits skewness or heteroscedasticity. It assumes that the response variable follows a distribution from the location-scale family, which includes a variety of specific distributions that cover most of the failure time distributions of engineering assets in industrial applications. Some examples of LLS distributions include {Normal}, {Log-Normal}, {Logistics}, {Log-Logistics}, {Smallest Extreme Value (SEV)}, and {Weibull}. Due to the versatility afforded by these distributions, LLS regression has been widely utilized in reliability engineering \citep{meeker2014statistical,doray1994ibnr,hong2010tool,hong2015bayesian,zhou2023federated} and industrial prognostics \citep{fang2017multistream,fang2019image,zhou2023supervised,fang2019Regression}. 

Similar to other statistical learning models, LLS regression needs a considerable amount of data for model training to achieve a desirable performance. However, the available training data related to failures from a single company/factory is usually insufficient to reliably train an LLS model. To address this challenge, companies may purchase a well-trained prognostic model from Original Equipment Manufacturers (OEMs). This is attributed to the fact that OEMs frequently lease a substantial number of equipment to customers and own the condition monitoring data of these leased equipment, providing them with sufficient failure-related data to train a dependable prognostic model. The trained models can be provided as a standalone product through subscription, ensuring that the model is regularly updated as additional failure-related data becomes available. The subscription allows OEMs to deliver reliable prognostic models to customers while safeguarding the privacy of their data—customers have access to the trained models without knowledge of the OEM's training data. However, recent research indicates that it is possible to infer information about the training data from the trained models by employing cyberattack techniques such as model inversion \citep{fredrikson2015model, wang2015regression, hidano2017model}. To tackle this challenge, research has shown that differential privacy is an effective technique for safeguarding data privacy \citep{wang2015regression, zhang2020broadening,abadi2016deep}. In alignment with this perspective, this article proposes privacy-preserving LLS regression based on differential privacy. 

Differential Privacy (DP)  is a foundational concept in privacy-preserving data analysis that aims to protect data privacy while extracting meaningful insights from datasets. It provides a rigorous mathematical framework for ensuring that the inclusion or exclusion of any data point does not unduly impact the outcome of a computation. Fundamentally, this is achieved by adding controlled noise to the results of statistical analyses. For example,  OEMs may employ DP to perturb estimated regression coefficients with random noise before transmitting them to customers. The amount of noise introduced in DP is determined by two factors: \textit{sensitivity} and \textit{privacy budget}. The first factor, sensitivity, is determined by the training data. It measures the maximum fluctuation in the output (e.g., the maximum change of regression coefficients) when a single data point is removed from or added to the training dataset. It's evident that to safeguard the privacy of training data, a larger noise is required when the sensitivity is higher. The second factor, privacy budget, is chosen by those who train the model. It controls the upper limit of the probability of potential information disclosure \citep{dwork2006calibrating}. This is necessary because, mathematically, DP (and many other privacy protection techniques) typically cannot guarantee a 100\% prevention of data leakage. Instead, they use a parameter (i.e., privacy budget) to control the probability that the data will not be leaked. This probability can be low (close to zero) or high. If a lower probability is sought, a higher degree of noise needs to be introduced into the model. Consequently, the performance of the model will be compromised more. In contrast, when the probability is high, the level of added noise is low, resulting in the performance of the DP-based model closely resembling that of the model without DP applied. Therefore, there exists a tradeoff between the probability of data leakage and the extent to which the model's performance is compromised.

In DP, the algorithm or procedure through which the noise is incorporated is referred to as \textit{mechanism}. One possible mechanism for regression analysis is to perturb the estimated regression coefficients with a random noise drawn from distributions like Laplace, Gaussian, or exponential. However, implementing this mechanism is often impractical because accurately determining the noise's amplitude relies on calculating the sensitivity, which usually is very challenging. To address this challenge, the commonly used mechanism in regression analysis is the \textit{functional mechanism}, which introduces noise to the objective function of the optimization problem linked to regression parameter estimation, rather than directly perturbing the regression coefficients themselves \citep{zhang2012functional}. In this article, the proposed differentially private LLS regression is based on the functional mechanism. The parameter estimation of LLS regression is typically achieved by using maximum likelihood estimation (MLE), which solves an optimization problem whose objective function is the log-likelihood function constructed using the training data and unknown regression parameters. To integrate differential privacy into LLS regression, we begin by decomposing the objective function (i.e., the log-likelihood function) through Taylor expansion, truncating at the second order. Then, we introduce noise to perturb the weights of the Taylor series. Subsequently, we reconstruct the perturbed objective function using the modified weights. Finally, we optimize the perturbed objective function to obtain the parameters of the LLS regression. We will derive the sensitivity of weights, which determines the magnitude of the noise introduced. Additionally, we will prove that the proposed regression method complies with differential privacy.

Commonly employed distributions in LLS regression include normal, log-normal, logistic, log-logistic, smallest extreme value (SEV), and Weibull. When the response variable is from a normal, logistic, or SEV distribution, the model is referred to as location-scale regression. In contrast, when the response variable is from a log-normal, log-logistic, or Weibull distribution, it is called log-location-scale regression. Location-scale regression and log-location-scale regression can be easily transformed into each other. For example, log-normal regression can be converted into normal regression by applying the logarithm to its response variable. Similarly, Weibull (or log-logistic) regression can be transformed into SEV (or logistic) regression by taking the logarithm of its response variable. Given that \cite{zhang2012functional} have explored differentially private normal regression, this article concentrates on the development of differentially private logistic regression and SEV regression (applicable to log-logistic regression and Weibull regression as well). It's important to clarify that the logistic regression investigated in this paper differs from logistic regression for classification \citep{wright1995logistic}. In this article, logistic regression refers to the response variable following a logistic distribution, a member of the location-scale family. Logistic regression for classification, on the other hand, involves a response variable from a binary or binomial distribution.

The remainder of this paper is organized as follows. Section 2 introduces the necessary preliminaries, encompassing LLS regression, differential privacy, and the functional mechanism. Section 3 outlines the proposed differentially private LLS regression. Following that, Sections 4 and 5 employ simulated and aircraft engine data from the NASA data repository to assess the performance of the proposed differentially private LLS regression. Finally, Section 6 provides concluding remarks.

\section{Preliminaries} \label{sec:preliminary}

In this section, we first review (log)-location-scale regression and then revisit some fundamental concepts of differential privacy. 

\subsection{LLS Regression}\label{sec:sec:lls}
Considering a dataset containing $n$ samples, we denote the data as $\{\tilde{y}_i,\{x_{ij}\}_{j=1}^d\}_{i=1}^n$, where $\tilde{y}_i$ is the dependent variable (i.e., response variable), which is corresponding to the time-to-failure (TTF) in reliability analysis and prognostics; $\{x_{ij}\}_{j=1}^d$ are the $d$-dimensional independent variables (i.e., predictors). Each $\tilde{y}_i$ is assumed to be sampled from a distribution from the (log)-location-scale family. In other words, $y_i$ is a random variable whose cumulative distribution function (CDF) is of the form 

\begin{equation}
    \operatorname{Pr}(y_i \leq t) =\Omega\left(\frac{t-\mu_i}{\sigma}\right),
\end{equation}

\noindent where $y_i=\tilde{y}_i$ if $\tilde{y}_i$ is from a distribution in the location-scale family and $y_i=\log(\tilde{y}_i)$ if $\tilde{y}_i$ is from a distribution in the log-location-scale family. $\Omega(\cdot)$ is the CDF of a standard distribution in the location-scale family. $\mu_i$ is referred to as the location parameter, and $\sigma$ is known as the scale parameter. Similar to the standard practices in regression analysis, it is assumed that the location parameter $\mu_i$ is determined by the independent variables, and the scale parameter $\sigma$ is not associated with the predictors. Denote $\mu_i=\beta_0+\sum_{j=1}^dx_{ij}\beta_j$, or $\mu_i=\sum_{j=0}^dx_{ij}\beta_j$ where $x_{i0}=1$ for simplicity, the CDF of $y_i$ can be expressed as follows

\begin{equation}
    \operatorname{Pr}(y_i \leq t) =\Omega\left(\frac{t-\sum_{j=0}^dx_{ij}\beta_j}{\sigma}\right),
\end{equation}

\noindent where $\{\beta_j\}_{j=0}^d$ are the unknown regression coefficients. These regression coefficients and the scale parameter $\sigma$ are usually estimated using maximum likelihood estimation (MLE). The likelihood function can be denoted as follows

\begin{equation}\label{eq:MLE}
L\left(\{\beta_j\}_{j=0}^d,\sigma\right)=\prod_{i=1}^n \frac{1}{\sigma} \omega\left(\frac{y_i-\sum_{j=0}^dx_{ij}\beta_j}{\sigma}\right)
\end{equation}

\noindent where $\omega(\cdot)$ is the probability density function (PDF) of a standard distribution in the location-scale family. For example, $\omega(z)=1/\sqrt{2\pi}\exp(-z^2/2)$ for normal distribution, $\omega(z)=\exp(z)/(1+\exp(z))^2$ for logistic distribution, and $\omega(z)=\exp(z)(z-\exp(z)))$ for SEV distribution. The estimation of $(\{\beta_j\}_{j=0}^d,\sigma)$ can be achieved by maximizing the log-likelihood function $\ell=\log(L(\{\beta_j\}_{j=0}^d,\sigma))$.

\subsection{Differential Privacy}

In differential privacy, a dataset can be seen as a collection of entries corresponding to individuals. For example, a dataset containing the TTF and predictors of $n$ machines (i.e., $\{\tilde{y}_i,\{x_{ij}\}_{j=1}^d\}_{i=1}^n$) can be seen as $n$ entries. Differential privacy guarantees that the modification (e.g., inclusion, exclusion, revision) of a single entry has no discernible statistical impact on the output. To define differential privacy, we first introduce the concept of neighboring datasets. Two datasets, $D$ and $D'$, are considered neighboring if they have the same number of entries while differing in one entry (corresponding to the data of one sample). With this concept, $\epsilon$-Differential Privacy is defined as follows.

\begin{definition}($\epsilon$-Differential Privacy \citep{dwork2006calibrating}). An algorithm $\mathcal{A}$ satisfies $\epsilon$-differential privacy, iff for any output $O$ of $\mathcal{A}$ and for any two neighbor databases $D$ and $D^{\prime}$, we have
\begin{equation}\label{eq:DP}
\operatorname{Pr}\left[\mathcal{A}\left(D\right)=O\right] \leq e^\epsilon \cdot \operatorname{Pr}\left[\mathcal{A}\left(D^{\prime}\right)=O\right]
\end{equation}
\end{definition}

\noindent Here, the operator $\operatorname{Pr}[\cdot]$ represents probability. $\mathcal{A}(\cdot)$ represents the mechanism, which denotes the method employed to introduce noise. Loosely speaking, $\operatorname{Pr}\left[\mathcal{A}\left(D\right)=O\right]$ means the probability that the output is $O$ using dataset $D$ and mechanism $\mathcal{A}(\cdot)$. Taking regression analysis using the functional mechanism as an example, $\operatorname{Pr}\left[\mathcal{A}\left(D\right)=O\right]$ means that the probability that the noisy weights are $O$ using training dataset $D$. Similarly, $\operatorname{Pr}\left[\mathcal{A}\left(D'\right)=O\right]$ means that the probability that the noisy weights are also $O$ using training dataset $D'$. Equation \eqref{eq:DP} quantifies the proximity of the probabilities to obtaining the same set of noisy weights when utilizing datasets $D$ and $D'$. The proximity is controlled by $e^\epsilon$, where $\epsilon$ is known as the \textit{privacy budget}. A smaller value of $\epsilon$ enforces a stronger privacy guarantee. For example, as $\epsilon$ approaches $0$, $e^{\epsilon}$ approaches $1$. As a result, the ratio between the two probabilities is closer to 1, and thus the two datasets become more indistinguishable.

In the aforementioned example of regression using the functional mechanism, noise is incorporated into the weights. However, a crucial aspect that has not been addressed is determining the magnitude of the noise. The magnitude of the noise is determined by the dataset. Mathematically, it is associated with global sensitivity, the definition of which is provided below.

\begin{definition}(Global Sensitivity \citep{dwork2006calibrating})
The global sensitivity of a function $f: D^n \rightarrow \boldsymbol{R}^d$
$$
G S_f(D)=\max _{D, D^{\prime}}\left\|f(D)-f\left(D^{\prime}\right)\right\|_1,
$$
\end{definition}

\noindent where $D$ and $D'$ are neighboring datasets. $f(\cdot)$ is a function that takes a dataset $D$ or $D'$ as the input, and the outputs are denoted as $f(D)$ and $f(D')$, respectively. In the functional mechanism, $f(D)$ and $f(D')$ are the values of the weights using dataset $D$ and $D'$, respectively. Thus, $G S_f(D)$ is the maximum changes that could occur in the weights when one sample in $D$ is replaced.

\section{Differentially Private LLS Regression} \label{s:methods}

In this section, we introduce differentially private (log)-location-scale regression. The commonly utilized distributions in the LLS family include normal, log-normal, logistic, log-logistic, SEV, and Weibull. Since \cite{zhang2012functional} have investigated differentially private normal regression, this article focuses on the development of differentially private logistic regression and SEV regression, which can be applied to log-logistic regression and Weibull regression as well. 

The proposed differentially private LLS regression relies on the functional mechanism, which works by perturbing the objective function used in the parameter estimation of LLS regression models. In other words, rather than solving the original log-likelihood function for parameter estimation, we optimize a perturbed version of it. 
This is achieved by first decomposing the objective function, specifically the log-likelihood function, through a Taylor expansion that is truncated at the second order. Then, noise is introduced to perturb the weights of the Taylor series. Subsequently, we reconstruct the perturbed objective function using the modified weights. Finally, optimization of the perturbed objective function is performed to obtain the parameters of the LLS regression. As mentioned earlier, the magnitude of the noise is directly linked to the sensitivity of the objective function. Thus, we will derive the sensitivity of the LLS regression models. Additionally, we will demonstrate that the proposed regression methods satisfy $\epsilon$-differential privacy.

Before constructing the proposed differentially private LLS regression models, we begin by standardizing the data. Following the notation in Section \ref{sec:sec:lls}, let's consider a dataset consisting of $n$ samples, denoted as $\{{y}_i,\{\tilde{x}_{ij}\}_{j=1}^d\}_{i=1}^n$, where $\{\tilde{x}_{ij}\}_{j=1}^d$ are the $d$-dimensional predictors and ${y}_i$ is the response (TTF or the logarithmic TTF). To scale the $j$th predictor, we use $x_{i j}=\frac{\tilde{x}_{i j}-\alpha_j}{\left(\beta_j-\alpha_j\right) \cdot \sqrt{d}}$, where $\alpha_j=\min\{\{\tilde{x}_{ij}\}_{i=1}^n\}$ and $\beta_j=\max\{\{\tilde{x}_{ij}\}_{i=1}^n\}$ denotes the minimum and maximum values of the $j$th predictor, respectively. This standardization ensures $\sqrt{\sum_{i=1}^d x_{i d}^2} \leq 1$, which will be employed to determine the global sensitivity of the proposed differentially private LLS regression. Additionally, we standardize the response variable such that $y_i \in [-1,1]$ for all $i$.

\subsection{Differentially Private SEV Regression}

Following the notations in the MLE function in Equation \eqref{eq:MLE}, the log-likelihood function of SEV regression can be expressed as follows
\begin{equation}
\ell(\{\beta_j\}_{j=0}^d,\sigma)=-n \log \sigma +\sum_{i=1}^n \frac{y_i-\sum_{j=0}^d \beta_j x_{i j}}{\sigma}-\sum_{i=1}^n \exp \left(\frac{y_i-\sum_{j=0}^d \beta_j x_{i j}}{\sigma}\right),
\quad 
\end{equation}

\noindent which is not concave. To make it a concave function, we apply the following transformation: $q = 1/\sigma$ and $p_j=\beta_j q$. As a result, the log-likelihood function can be re-expressed as 
\begin{equation}\label{eq:llk}
\ell(\{p_j\}_{j=0}^d,q)=n \log q +\sum_{i=1}^n \left(y_iq-\sum_{j=0}^d p_j x_{i j}\right)-\sum_{i=1}^n \exp \left(y_iq-\sum_{j=0}^d p_j x_{i j}\right),
\quad 
\end{equation}

To integrate differential privacy with SEV regression, as shown in Proposition \ref{prop:sevtaylor}, we first employ Taylor expansion to expand the log-likelihood function in Equation \eqref{eq:llk} as a weighted combination of polynomial functions of $\{p_j\}_{j=0}^d$ and $q$.

\begin{proposition}\label{prop:sevtaylor}
By applying Taylor expansion and truncating it at the second order, Equation \eqref{eq:llk} can be re-written as follows
\begin{multline}\label{eq:sevtaylor}
 \tilde{\ell}\left(\{p_j\}_{j=0}^d,q\right) =    - \frac{5n}{2} + 2nq- \frac{1}{2}\left(n+ \sum_{i=1}^n y_i^2\right) q^2   + \sum_{i=1}^n\sum_{j=0}^d y_i x_{i j} p_j q - \frac{1}{2} \sum_{i=1}^n \sum_{j=0}^d x_{i j}^2 p_j^2 \\-\frac{1}{2}\sum_{i=1}^n \left(\sum_{h=0, h\neq j}^d \sum_{j=0}^d x_{i j} x_{i h} p_j p_h \right),
\end{multline}
\end{proposition} 

\noindent The proof is available in the Appendix. Let $w_1 = - \frac{5n}{2}, w_q=2n, w_{q^2} = - \frac{1}{2}\left(n+ \sum_{i=1}^n y_i^2\right)$, $\{w_{p_jq}=\sum_{i=1}^n y_i x_{i j}\}_{j=0}^d,\{w_{p_j^2}=- \frac{1}{2} \sum_{i=1}^n x_{i j}^2\}_{j=0}^d$, and $\{\{w_{p_jp_h}=-\frac{1}{2}\sum_{i=1}^n x_{i j} x_{i h}\}_{j=0}^d\}_{h=0,h\neq j}^d$, Equation \eqref{eq:sevtaylor} can be rewritten as $\tilde{\ell}\left(\{p_j\}_{j=0}^d,q\right) =  w_1 + w_qq +w_{q^2} q^2  +\sum_{j=0}^d w_{p_jq} p_j q +  \sum_{j=0}^d w_{p_j^2}p_j^2 + \sum_{h=0, h\neq j}^d \sum_{j=0}^d w_{p_jp_h} p_j p_h $. To incorporate differential privacy into SEV regression, we then introduce noise to perturb the weights. As discussed earlier, the magnitude of the noise is determined by the sensitivity and the privacy budget. Proposition \ref{pro:sevnoise} provides an upper bound for the sensitivity of the polynomial coefficients of the function $\tilde{\ell}\left(\{p_j\}_{j=0}^d,q\right)$ in Equation \eqref{eq:sevtaylor}.

\begin{proposition}\label{pro:sevnoise}
Let $D$ and $D^{\prime}$ be any two neighbor databases, and let $\tilde{\ell}_D\left(\left\{p_j\right\}_{j=0}^d, q\right)$ and $\tilde{\ell}_{D^{\prime}}\left(\left\{p_j\right\}_{j=0}^d, q\right)$ be the objective functions on $D$ and $D^{\prime}$, respectively. Also, let the weights of $\tilde{\ell}_D\left(\left\{p_j\right\}_{j=0}^d, q\right)$ be $w_1^D, w_q^D,w_{q^2}^D,\{w_{p_jq}^D\}_{j=1}^d,\{w_{p_j^2}^D\}_{j=0}^d,\{\{w_{p_jp_h}^D\}_{j=0}^d\}_{h=0,h\neq j}^d$ and the weights of $\tilde{\ell}_{D'}\left(\left\{p_j\right\}, q\right)$ be $w_1^{D'}, w_q^{D'},w_{q^2}^{D'},\{w_{p_jq}^{D'}\}_{j=0}^d,\{w_{p_j^2}^{D'}\}_{j=0}^d,\{\{w_{p_jp_h}^{D'}\}_{j=0}^d\}_{h=0,h\neq j}^d$. Then, we have the following inequality:
\begin{equation}
\begin{split}
       \|w_1^D-w_1^{D'}\|_1 + \|w_q^D-w_q^{D'}\|_1 + \|w_{q^2}^D-w_{q^2}^{D'}\|_1 + \sum_{j=0}^d\|w_{p_jq}^D-w_{p_jq}^{D'}\|_1 \\+ \sum_{j=0}^d\|w_{p_j^2}^D-w_{p_j^2}^{D'}\|_1+\sum_{j=0}^d\sum_{h=0,h\neq j}^d\|w_{p_jp_h}^D-w_{p_jp_h}^{D'}\|_1\leq 4+4\sqrt{d}+d.
\end{split}
\end{equation}

\end{proposition}

\noindent The proof of Proposition \ref{pro:sevnoise} can be found in the Appendix. Based on the inequality provided by Proposition \ref{pro:sevnoise} and let $\Delta=4+4\sqrt{d}+d$, we perturb each weight in Equation \eqref{eq:sevtaylor} by adding $\operatorname{Lap}(\Delta/\epsilon)$ to it, where $\operatorname{Lap}(\Delta/\epsilon)$ represents a random variable drawn from the Laplace distribution with zero mean and scale $\Delta/\epsilon$, and $\epsilon$ is the privacy budget. We denote the perturbed weights as  $\tilde{w}_1, \tilde{w}_q, \tilde{w}_{q^2}$, $\{\tilde{w}_{p_jq}\}_{j=0}^d,\{\tilde{w}_{p_j^2}\}_{j=0}^d$, and $\{\{\tilde{w}_{p_jp_h}\}_{j=0}^d\}_{h=0,h\neq j}^d$, where, for instance, $\tilde{w}_1=w_1+\operatorname{Lap}\left(\Delta/\epsilon\right)$. As a result, the perturbed objective function for the parameter estimation of SEV regression can be denoted as follows:
\begin{equation}\label{eq:sevobj}
\tilde{\tilde{\ell}}\left(\{p_j\}_{j=0}^d,q\right) =  \tilde{w}_1 + \tilde{w}_q +\tilde{w}_{q^2} q^2  +\sum_{j=0}^d \tilde{w}_{p_jq} p_j q +  \sum_{j=0}^d \tilde{w}_{p_j^2}p_j^2 + \sum_{j=0}^d \sum_{h=0,h\neq j}^d \tilde{w}_{p_jp_h} p_j p_h.
\end{equation}

\noindent Optimizing $\max_{\{p_j\}_{j=0}^d,q}\tilde{\tilde{\ell}}\left(\{p_j\}_{j=0}^d,q\right)$ yields the estimated parameters $\{\{\hat{p}_j\}_{j=0}^d,\hat{q}\}$, which can be easily transformed back to the regression parameters in SEV regression: $\hat{\sigma}=1/\hat{q}$ and $\hat{\beta}_j=\hat{p}_j\hat{\sigma}$. We summarize the parameter estimation algorithm for the differentially private SEV regression using the functional mechanism in Algorithm 1.

\begin{algorithm}[H]
\begin{algorithmic}[1]
\REQUIRE Database $\{\tilde{y}_i,\{x_{ij}\}_{j=0}^d\}_{i=1}^n$, privacy budget $\epsilon$
\ENSURE $\epsilon$-DP protected parameter estimates $\hat{\sigma},\{\hat{\beta}_j\}_{j=0}^d$
\STATE Set $\Delta=4+4\sqrt{d}+d$

\STATE $\tilde{w}_1=w_1+\operatorname{Lap}\left(\Delta/\epsilon\right)$, $\tilde{w}_q=w_q+\operatorname{Lap}\left(\Delta/\epsilon\right)$, $\tilde{w}_{q^2}=w_{q^2}+\operatorname{Lap}\left(\Delta/\epsilon\right)$\\
\STATE $\{\tilde{w}_{p_jq}={w}_{p_jq}+\operatorname{Lap}\left(\Delta/\epsilon\right)\}_{j=0}^d$\\
\STATE $\{\tilde{w}_{p_j^2}={w}_{p_j^2}+\operatorname{Lap}\left(\Delta/\epsilon\right)\}_{j=0}^d$\\
\STATE $\{\{\tilde{w}_{p_jp_h}={w}_{p_jp_h}+\operatorname{Lap}\left(\Delta/\epsilon\right)\}_{j=0}^d\}_{h=0,h\neq j}^d$
\STATE Solving $\{\{\hat{p}_j\}_{j=0}^d,\hat{q}\}=\max_{\{p_j\}_{j=0}^d,q}\tilde{\tilde{\ell}}\left(\{p_j\}_{j=0}^d,q\right)$, where $\tilde{\tilde{\ell}}\left(\{p_j\}_{j=0}^d,q\right)$ is defined in Equation \eqref{eq:sevobj}
\STATE  Set $\hat{\sigma}=1/\hat{q}, \{\hat{\beta}_j=\hat{p}_j\hat{\sigma}\}_{j=0}^d$.
\STATE Return $\hat{\sigma},\{\hat{\beta}_j\}_{j=0}^d$
\end{algorithmic}
\caption{\textbf{Parameter Estimation for Differentially Private SEV Regression} }\label{alg:sev}
\end{algorithm}

Proposition \ref{prop:alg1} shows that Algorithm 1 ensures $\epsilon$-differential privacy, with the proof available in the Appendix.

\begin{proposition}\label{prop:alg1}
Algorithm 1 satisfies $\epsilon$-differential privacy.
\end{proposition}

\subsection{Differentially Private Logistic Regression}

The log-likelihood function of logistic regression can be expressed as follows

\begin{equation}
\ell(\{\beta_j\}_{j=0}^d,\sigma)=-n \log \sigma+\sum_{i=1}^n \frac{y_i-\sum_{j=0}^d \beta_j x_{i j}}{\sigma}-2 \sum_{n=1}^n \log \left(1+\exp\left(\frac{y_i-\sum_{j=0}^d \beta_j x_{i j}}{\sigma}\right)\right).
\end{equation}

\noindent which is also not concave. Thus, the following transformation is applied to make it concave $q = 1/\sigma$ and $p_j=\beta_j q$. As a result, the log-likelihood function can be re-expressed as 
\begin{equation}\label{eq:lgllk}
\ell(\{p_j\}_{j=0}^d,q)=n \log q+\sum_{i=1}^n \left(y_iq-\sum_{j=0}^d p_j x_{i j}\right)-2 \sum_{i=1}^n \log \left(1+\exp\left(y_iq-\sum_{j=0}^d p_j x_{i j}\right)\right).
\end{equation}

Similar to SEV regression, to integrate differential privacy into the logistic regression, we first express the function $\ell({p_j}{j=0}^d, q)$ in Equation \eqref{eq:lgllk} as a polynomial function of $\{p_j\}_{j=0}^d$ and $q$. This expansion is given in Proposition \ref{prop:lgtaylor}.

\begin{proposition}\label{prop:lgtaylor}
By applying Taylor expansion and truncating it at the second order, Equation \eqref{eq:lgllk} can be rewritten as 
\begin{multline}\label{eq:lgtaylor}
\tilde{\ell}\left(\{p_j\}_{j=0}^d,q\right) = -n \left(\frac{3}{2}+ 2\log 2\right) + 2nq - \left(\frac{n}{2}+\frac{1}{4} \sum_{i=1}^n y_i^2 \right) q^2  + \frac{1}{2} \sum_{i=1}^n\sum_{j=0}^d y_i x_{i j} p_j q \\
- \frac{1}{4} \sum_{i=1}^n \sum_{j=0}^d x_{i j}^2 p_j^2 - \frac{1}{4}\sum_{i=1}^n \left(\sum_{h=0, h\neq j}^d \sum_{j=0}^d x_{i j} x_{i h} p_j p_h \right).
\end{multline}

\end{proposition}

\noindent The proof of Proposition \ref{prop:lgtaylor} can be found in the Appendix. Denoting $w_1 =  -n \left(\frac{3}{2}+ 2\log 2\right)$, $w_q=2n$, $w_{q^2} =-\left(\frac{n}{2}+\frac{1}{4} \sum_{i=1}^n y_i^2\right)$, $\{w_{p_jq}=\frac{1}{2} \sum_{i=1}^ny_i x_{i j}\}_{j=0}^d$, $\{w_{p_j^2}=-\frac{1}{4} \sum_{i=1}^n x_{i j}^2\}_{j=0}^d$, $\{\{w_{p_jp_h}=- \frac{1}{4}\sum_{i=1}^n x_{i j} x_{i h}\}_{j=0}^d\}_{h=0,h\neq j}^d$, Equation \eqref{eq:lgtaylor} can be rewritten as $\tilde{\ell}\left(\{p_j\}_{j=0}^d,q\right) = w_1 + w_q q +w_{q^2} q^2 + \sum_{j=0}^d w_{p_jq}p_j q + \sum_{j=0}^d w_{p_j^2} p_j^2 + \sum_{j=0}^d \sum_{h=0,h\neq j}^d w_{p_jp_h} p_j p_h $. To integrate differential privacy with logistic regression, we then introduce noise to the weights. To determine the amplitude of noise, we first provide an upper bound for the sensitivity of the polynomial coefficients of the function in Equation \eqref{eq:lgtaylor} in Proposition \ref{prop:lgsensitiviy} below. 

\begin{proposition}\label{prop:lgsensitiviy}

Let $D$ and $D^{\prime}$ be any two neighbor databases, and let $\tilde{\ell}_D\left(\left\{p_j\right\}_{j=0}^d, q\right)$ and $\tilde{\ell}_{D^{\prime}}\left(\left\{p_j\right\}_{j=0}^d, q\right)$ be the objective functions on $D$ and $D^{\prime}$, respectively. Also, let the weights of $\tilde{\ell}_D\left(\left\{p_j\right\}_{j=0}^d, q\right)$ be $w_1^D, w_q^D,w_{q^2}^D,\{w_{p_jq}^D\}_{j=1}^d,\{w_{p_j^2}^D\}_{j=0}^d,\{\{w_{p_jp_h}^D\}_{j=0}^d\}_{h=0,h\neq j}^d$ and the weights of $\tilde{\ell}_{D'}\left(\left\{p_j\right\}, q\right)$ be $w_1^{D'}, w_q^{D'},w_{q^2}^{D'},\{w_{p_jq}^{D'}\}_{j=0}^d,\{w_{p_j^2}^{D'}\}_{j=0}^d,\{\{w_{p_jp_h}^{D'}\}_{j=0}^d\}_{h=0,h\neq j}^d$. Then, we have the following inequality:
\begin{equation}
\begin{split}
       \|w_1^D-w_1^{D'}\|_1 + \|w_q^D-w_q^{D'}\|_1 + \|w_{q^2}^D-w_{q^2}^{D'}\|_1 + \sum_{j=0}^d\|w_{p_jq}^D-w_{p_jq}^{D'}\|_1 + \sum_{j=0}^d\|w_{p_j^2}^D-w_{p_j^2}^{D'}\|_1\\+\sum_{j=0}^d\sum_{h=0,h\neq j}^d\|w_{p_jp_h}^D-w_{p_jp_h}^{D'}\|_1\leq 2 + 2\sqrt{d} + \frac{1}{2}d.
\end{split}
\end{equation}
\end{proposition}

The proof of Proposition \ref{prop:lgsensitiviy} is provided in the Appendix. According to Proposition \ref{prop:lgsensitiviy}, each weight in Equation \eqref{eq:lgtaylor} can be perturbed by adding a random value drawn from the Laplace distribution with zero mean and scale $\Delta/\epsilon$, where $\epsilon$ is the privacy budget, and $\Delta = 2 + 2\sqrt{d} + \frac{1}{2}d$. If we denote the perturbed weights as  $\tilde{w}_1, \tilde{w}_q, \tilde{w}_{q^2}$, $\{\tilde{w}_{p_jq}\}_{j=0}^d,\{\tilde{w}_{p_j^2}\}_{j=0}^d$, and $\{\{\tilde{w}_{p_jp_h}\}_{j=0}^d\}_{h=0,h\neq j}^d$, then $\tilde{w}_1=w_1+\operatorname{Lap}\left(\Delta/\epsilon\right)$, $\tilde{w}_q=w_q+\operatorname{Lap}\left(\Delta/\epsilon\right),\cdots$. As a result, the new objective function for the parameter estimation of logistic regression can be denoted as follows:

\begin{equation}\label{eq:lgobj}
\tilde{\tilde{\ell}}\left(\{p_j\}_{j=0}^d,q\right) =  \tilde{w}_1 + \tilde{w}_q +\tilde{w}_{q^2} q^2  +\sum_{j=0}^d \tilde{w}_{p_jq} p_j q +  \sum_{j=0}^d \tilde{w}_{p_j^2}p_j^2 + \sum_{j=0}^d \sum_{h=0,h\neq j}^d \tilde{w}_{p_jp_h} p_j p_h.
\end{equation}

\noindent Solving the optimization problem $\max_{\{p_j\}_{j=0}^d,q}\tilde{\tilde{\ell}}\left(\{p_j\}_{j=0}^d,q\right)$ yields the estimated parameters $\{\{\hat{p}_j\}_{j=0}^d,\hat{q}\}$, which can then be transformed back to the regression parameters in logistic regression: $\hat{\sigma}=1/\hat{q}$ and $\hat{\beta}_j=\hat{p}_j\hat{\sigma}$. We summarize the parameter estimation algorithm for the differentially private logistic regression using the functional mechanism in Algorithm 2.

\begin{algorithm}[H]
\begin{algorithmic}[1]
\REQUIRE Database $\{\tilde{y}_i,\{x_{ij}\}_{j=0}^d\}_{i=1}^n$, privacy budget $\epsilon$
\ENSURE $\epsilon$-DP protected parameter estimates $\hat{\sigma},\{\hat{\beta}_j\}_{j=0}^d$
\STATE Set $\Delta=2 + 2\sqrt{d} + \frac{1}{2}d$

\STATE $\tilde{w}_1=w_1+\operatorname{Lap}\left(\Delta/\epsilon\right)$, $\tilde{w}_q=w_q+\operatorname{Lap}\left(\Delta/\epsilon\right)$, $\tilde{w}_{q^2}=w_{q^2}+\operatorname{Lap}\left(\Delta/\epsilon\right)$\\
\STATE $\{\tilde{w}_{p_jq}={w}_{p_jq}+\operatorname{Lap}\left(\Delta/\epsilon\right)\}_{j=0}^d$\\
\STATE $\{\tilde{w}_{p_j^2}={w}_{p_j^2}+\operatorname{Lap}\left(\Delta/\epsilon\right)\}_{j=0}^d$\\
\STATE $\{\{\tilde{w}_{p_jp_h}={w}_{p_jp_h}+\operatorname{Lap}\left(\Delta/\epsilon\right)\}_{j=0}^d\}_{h=0,h\neq j}^d$
\STATE Solving $\{\{\hat{p}_j\}_{j=0}^d,\hat{q}\}=\max_{\{p_j\}_{j=0}^d,q}\tilde{\tilde{\ell}}\left(\{p_j\}_{j=0}^d,q\right)$, where $\tilde{\tilde{\ell}}\left(\{p_j\}_{j=0}^d,q\right)$ is defined in Equation \eqref{eq:lgobj}
\STATE  Set $\hat{\sigma}=1/\hat{q}, \{\hat{\beta}_j=\hat{p}_j\hat{\sigma}\}_{j=0}^d$.
\STATE Return $\hat{\sigma},\{\hat{\beta}_j\}_{j=0}^d$
\end{algorithmic}
\caption{\textbf{Parameter Estimation for Differentially Private Logistic Regression} }\label{alg:log}
\end{algorithm}

Proposition \ref{prop:lgalg} indicates that Algorithm 2 ensures $\epsilon$-differential privacy, and the proof can be found in the Appendix.

\begin{proposition}\label{prop:lgalg}
   Algorithm \ref{alg:log} satisfies $\epsilon$-differential privacy. 
\end{proposition}

\section{Simulation Study} \label{sec:sim}

In this section, we conduct simulation studies to evaluate the performance of the proposed differentially private LLS regression models.

\subsection{Experimental settings} 

We generate data for $n$ samples (the specific values of $n$ will be discussed later). First, each entry of the predictors of sample $i$, $\boldsymbol{x}_i=(x_{i1}, x_{i2}, \ldots, x_{id})^\top$, is randomly drawn from a standard normal distribution, $\mathcal{N}(0,1)$. Here, $d$ is the dimension of features, whose specific values will be discussed later as well. Next, we generate the regression coefficients $\boldsymbol {\beta}=(\beta_0,\beta_1,\ldots,\beta_d)^\top$, each element of which is generated from a standard normal distribution, $\mathcal{N}(0,1)$. Third, the response of sample $i$ is generated using $\tilde{y}_i=(1, \boldsymbol{x}_i^\top)\boldsymbol{\beta}+\varepsilon_i$, where $\varepsilon_i$ is a random number sampled from either an SEV or logistic distribution (depending on the regression model being evaluated), with the location parameter set to 0 and the scale parameter set to 1.

We use the simulated data to assess the performance of the proposed differentially private LLS regression, denoted as ``DP." The benchmark employed in this study is the classical LLS regression, designated as ``Non-DP." Our interest lies in investigating how the proposed method performs under different conditions, including (1) predictor dimensions (i.e., $d$), (2) training sample sizes (i.e., $n$), and (3) privacy budgets (i.e., $\epsilon$). These factors are expected to influence the performance of the proposed DP method. We summarize the specific settings of these three factors used in this simulation study as follows. 

\begin{itemize}
    
    \item To examine the impact of predictor dimension on the performance of the proposed DP LLS regression, the tested values for $d$ are $20$, $22$, $24$, $26$, $28$, $30$, $32$, $34$, $36$ and $38$. For SEV regression, we set $n=1e4$. For logistic regression, $n=5e3$. We have $\epsilon=0.5$ for both models.
    
    \item To explore the influence of sample size, the values of $n$ to be tested are $1e4$, $1.5e4$, $2e4$, $2.5e4$, $3e4$, $3.5e4$, $4e4$, $4.5e4$, $5e4$, $5.5e4$, and $6e4$ for SEV regression. For logistic regression, the tested $n$ values are $5e3$, $6e3$, $7e3$, $8e3$, $9e3$, $1e4$, $2e4$, $3e4$, $4e4$, $5e4$ and $6e4$. We have $d = 35$ and $\epsilon=0.5$ for both models.
    
    \item To assess the influence of the privacy budget, the tested values of $\epsilon$ are $0.3$, $0.4$, $0.5$, $0.6$, $0.7$, $0.8$, $0.9$, $1$, and $2$, while we set $d=25$ for SEV regression and $d=38$ for logistic regression. For both regression models, we set $n=1e4$.
\end{itemize}

For each of the trials above, we randomly select $80\%$ of the data for model training. The remaining $20\%$ data are used for testing. The metric used to evaluate the performance is the absolute prediction error, which is computed using $|(\hat{\tilde{y}}_i-\tilde{y}_i)/\tilde{y}_i|$, where $\hat{\tilde{y}}_i$ and $\tilde{y}_i$ are the predicted and true response of sample $i$, respectively. Each trial will be repeated 100 times, by setting the random seed as the repetition number. The prediction errors of the 100 repetitions are reported in a single boxplot.

\subsection{Results and Analysis}

\subsubsection{Prediction Error vs. Predictor Dimension}

In the initial set of experiments, we investigate the performance of DP and non-DP models across various levels of feature dimensions. The dimensionality is crucial as it determines the sensitivity, influencing the amplitude of noise introduced to the objective function of parameter estimation. As discussed in Section 2, we introduce noise using \(\operatorname{Lap}\left((4 + 4\sqrt{d} + d)/\epsilon\right)\) for the SEV regression model and \(\operatorname{Lap}\left((2 + 2\sqrt{d} + \frac{1}{2}d)/\epsilon\right)\) for the logistic model, where $d$ represents the feature dimension. Keeping the sample size constant at \(1e4\) for SEV regression and \(5e3\) for logistic regression, and setting the privacy budget to \(\epsilon = 0.5\). For both regressions, we conducted tests for $d$ values ranging from 20 to 38, in increments of 2. We present the prediction errors for SEV regression in Figure 1 and for Logistic regression in Figure 2. 

\begin{figure}[!h]
\centering
\includegraphics[scale=0.5]{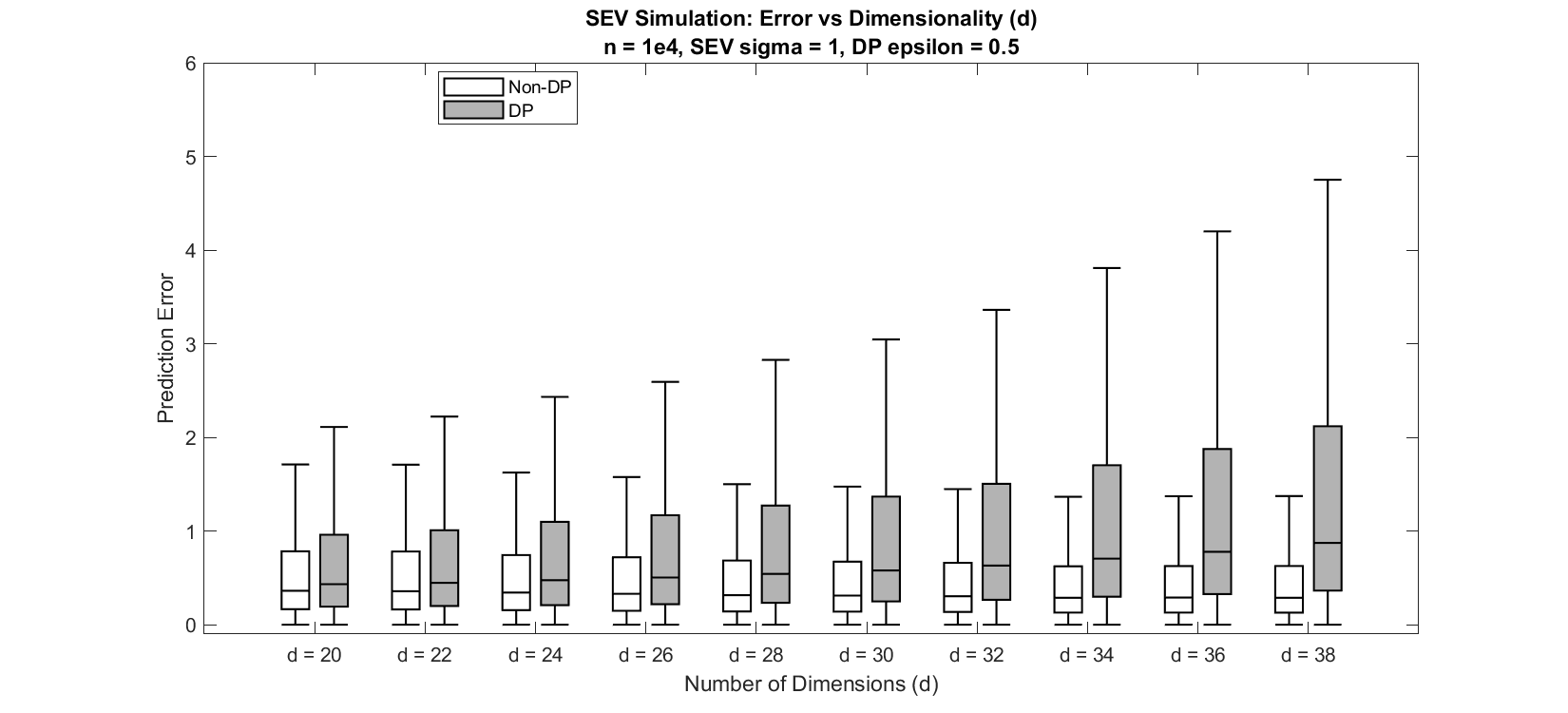}
\caption{SEV Regression: Prediction Error vs. Dimensionality $d$}
\end{figure}

\begin{figure}[!h]
\centering
\includegraphics[scale=0.5]{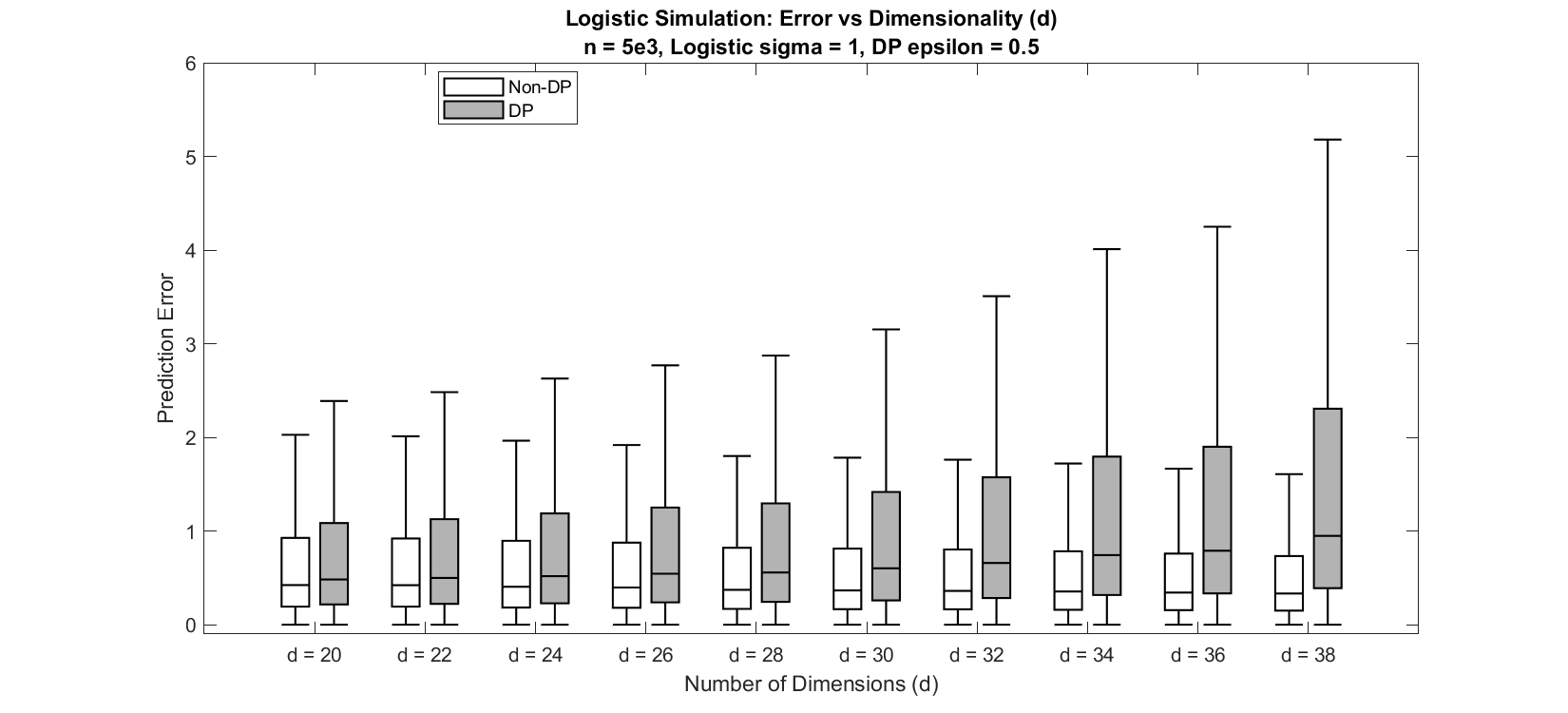}
\caption{Logistic Regression: Prediction Error vs. Dimensionality $d$}
\end{figure}

In Figures 1 and 2, the horizontal axis is the dimension of predictors, and the vertical axis is the prediction error. For each predictor dimension, the left boxplot represents the non-DP model, while the right one corresponds to the DP model. Figures 1 and 2 reveal that the prediction errors of the proposed DP models deteriorate with the increase in feature dimension, whereas the prediction errors of the non-DP models remain consistent with the rise in feature dimension. For instance, when \(d=20\), the median errors for both non-DP and DP models are approximately \(0.45\) for Logistic regression. However, as \(d\) expands to \(26\), the median error for the DP model increases to \(0.54\), while the non-DP model maintains a stable error of \(0.40\). This trend persists with further increases in dimensionality. At \(d=32\), the median error in the DP model rises to \(0.66\), and at \(d=36\), it further grows to \(0.79\). In contrast, the errors in the non-DP model remain relatively unaffected by the changes in \(d\), emphasizing the widening gap in prediction errors between DP and non-DP models as the feature dimensionality increases. This increase in prediction errors of DP models is anticipated and can be attributed to the escalating magnitude of Laplace noise added to the objective function for parameter estimation. Recall that the noise added to the SEV model is characterized by a Laplace distribution with a scale parameter $((4 + 4\sqrt{d} + d)/\epsilon)$, and the noise for the logistic model follows a Laplace distribution with a scale parameter $((2 + 2\sqrt{d} + \frac{1}{2}d)/\epsilon)$. In both cases, the magnitude of the noise increases with the feature dimension $d$. 

Figures 1 and 2 also illustrate that the performance of the DP models remains comparable to that of the non-DP models when the dimension of features is smaller than or equal to 25, using the given fixed sample size. This suggests that it is feasible to safeguard data privacy using DP without significantly compromising the performance of the LLS regression models, particularly when the feature dimension is relatively small compared to the sample size. In real-world applications, it is beneficial to employ some dimension reduction or variable selection techniques to reduce the number of predictors before applying the differentially private LLS regression.

\subsubsection{Prediction Error vs. Training Sample Size}

In the following experiments, we investigate the correlation between prediction error and the size of training samples. For this purpose, we fix the feature dimension $d = 35$, and the privacy budget, \(\epsilon\), fixed at \(0.5\). The sample sizes investigated ranges from \(5 \times 10^3\) to \(6 \times 10^4\), including various selected values for SEV and logistic regressions. 

Figures 3 and 4 present the error of experiments with the increase in training sample size. It can be observed that the performance of non-DP models remains relatively stable, whereas DP models show improved performance that increasingly aligns with that of non-DP models as the dataset size grows. For example, in the results for the SEV regression, the median error for the DP model is around \(0.76\) at \(n = 1 \times 10^4\), while it is only \(0.29\) for the non-DP model. However, this discrepancy diminishes as \(n\) grows. At \(n = 3 \times 10^4\), the median error for the DP model decreases to \(0.35\), while the non-DP model stands at \(0.28\), and it continues to decrease to \(0.29\) at \(n = 6 \times 10^4\), with the non-DP model's error at \(0.27\). A similar pattern is observed in the results of the logistic regression. This is reasonable since a larger sample size enhances the representation of the underlying population, naturally reducing prediction error in regression models. 

\begin{figure}[!h]
\centering
\includegraphics[scale=0.5]{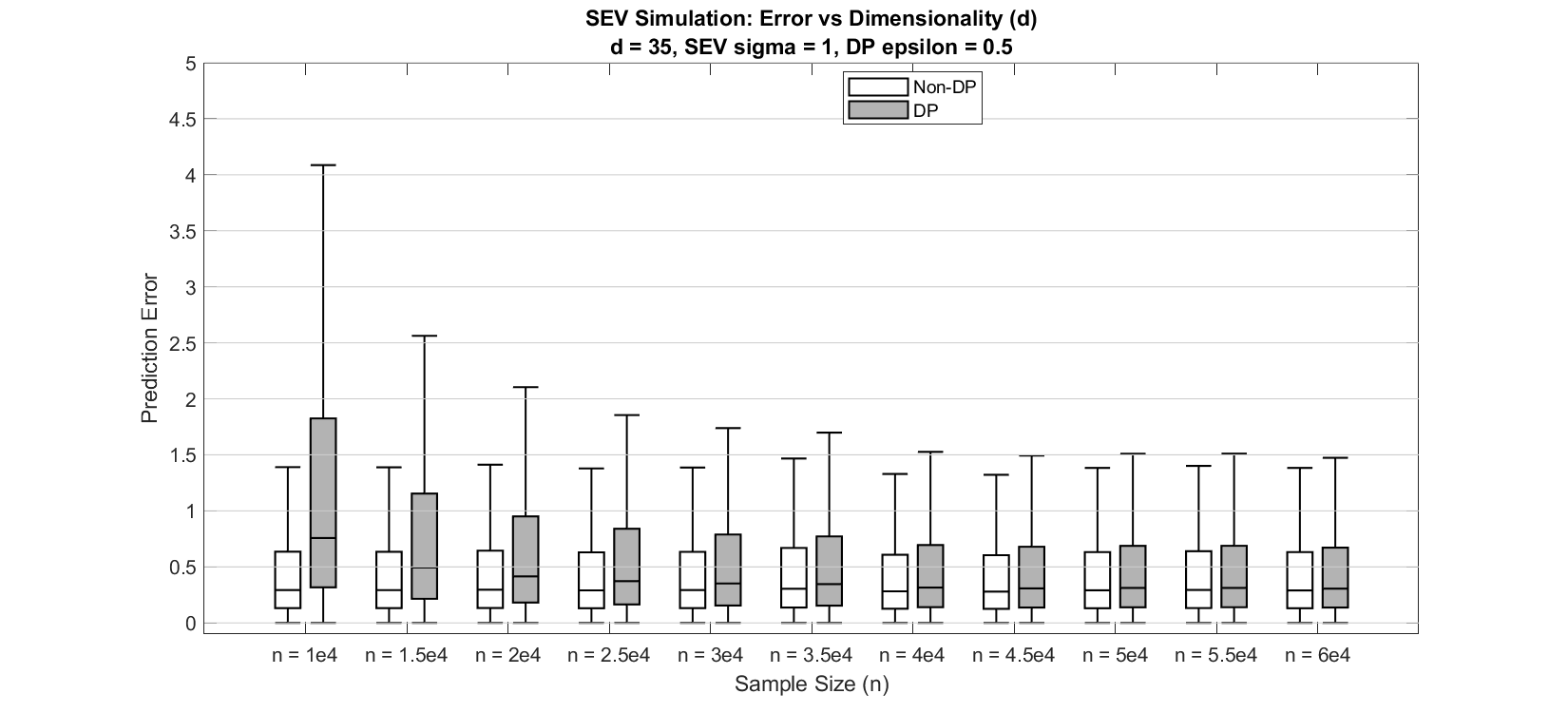}
\caption{SEV Regression: Error vs Sample Size $n$}
\end{figure}

\begin{figure}[!h]
\centering
\includegraphics[scale=0.5]{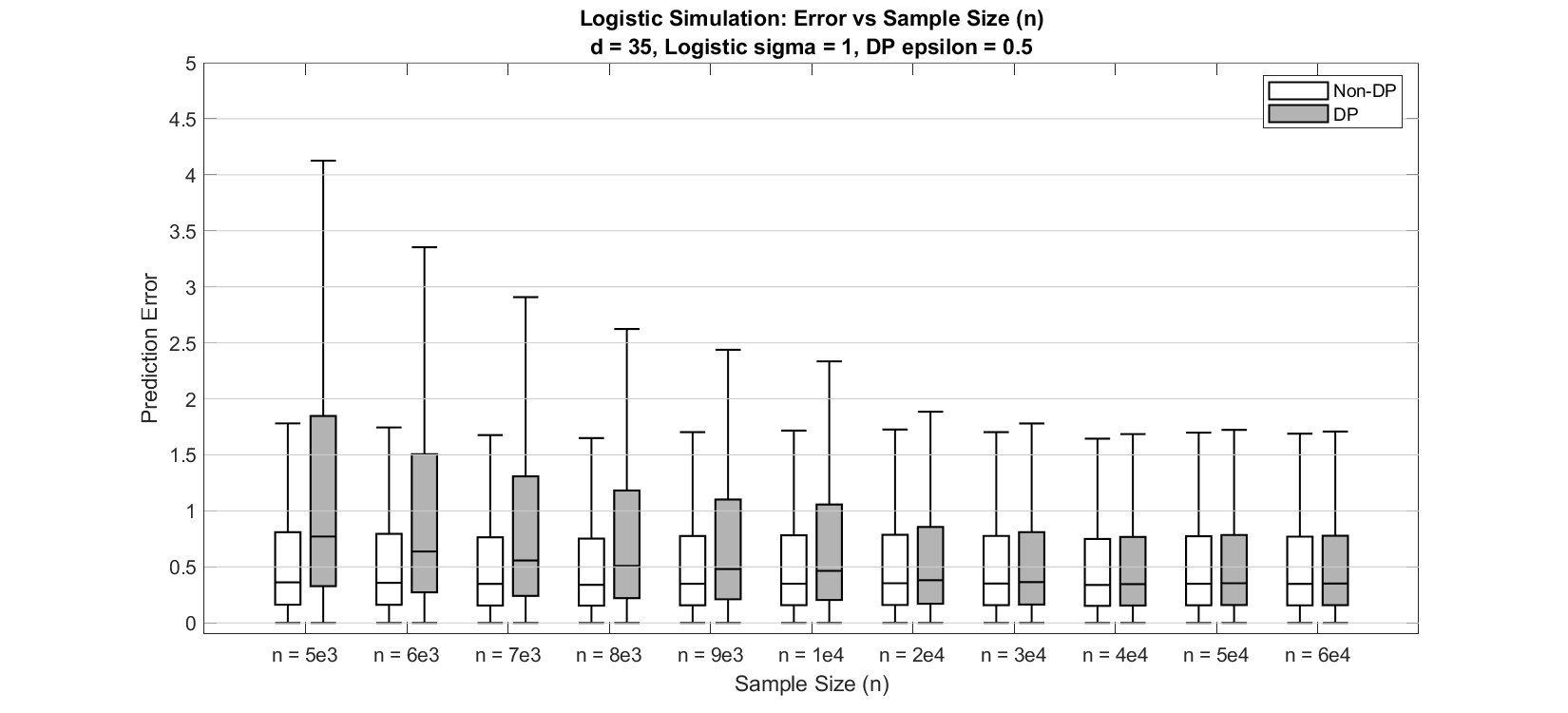}
\caption{Logistic Regression: Error vs Sample Size $n$}
\end{figure}

In this set of experiments, we utilized a relatively large dimension number, $35$, and based on the results, we can infer that a sample size of $4\times 10^4$ and $2\times 10^4$ is required for the proposed DP models to achieve results comparable to the non-DP models in terms of SEV and logistic regressions, respectively. Furthermore, we observed nearly identical error levels when the sample size reached \(6 \times 10^4\) for SEV regression and \(4 \times 10^4\) for logistic regression. In other words, the performance converged and did not continue to improve with further increases in sample size beyond it. It is expected that, with a smaller value of feature dimension \(d\), the convergence requirements of \(n\) would become even more accommodating. 
Additionally, we observed that the non-DP models did not exhibit the same decreasing pattern of prediction error as the sample size increased within our experimental range. We attribute this to the fact that the smallest sample size in this study is  $5\times 10^3$, which is already sufficiently large for the feature dimension of $35$, indicating that the prediction performance has already converged.

\subsubsection{Prediction Error vs. Privacy Budget}
In the last set of experiments, our focus is on examining the impact of the privacy budget on the performance of the proposed LLS regression. This offers insights into the trade-offs between the level of privacy protection and prediction accuracy. Here, we conduct experiments across a range of \(\epsilon\) values including \(0.3, 0.4, 0.5, 0.6, 0.7, 0.8, 0.9, 1\), and \(2\), and fix the feature dimension at \(d = 25\) for SEV regression, at $d=38$ for logistic regression. The sample size is set as \(n = 1 \times 10^4\) for both regressions.

\begin{figure}[!h]
\centering
\includegraphics[scale=0.5]{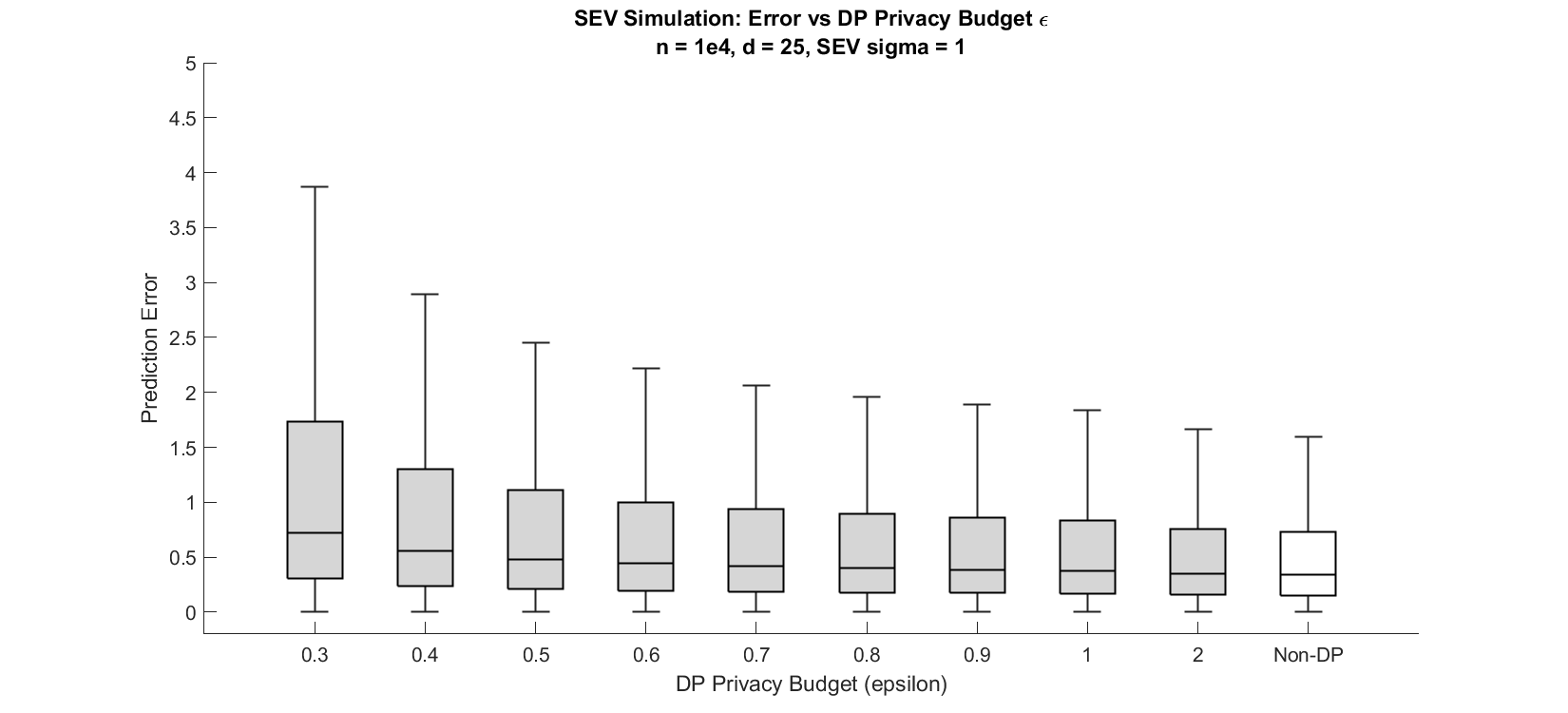}
\caption{SEV Regression: Error vs DP privacy budget $\epsilon$}
\end{figure}

\begin{figure}[!h]
\centering
\includegraphics[scale=0.5]{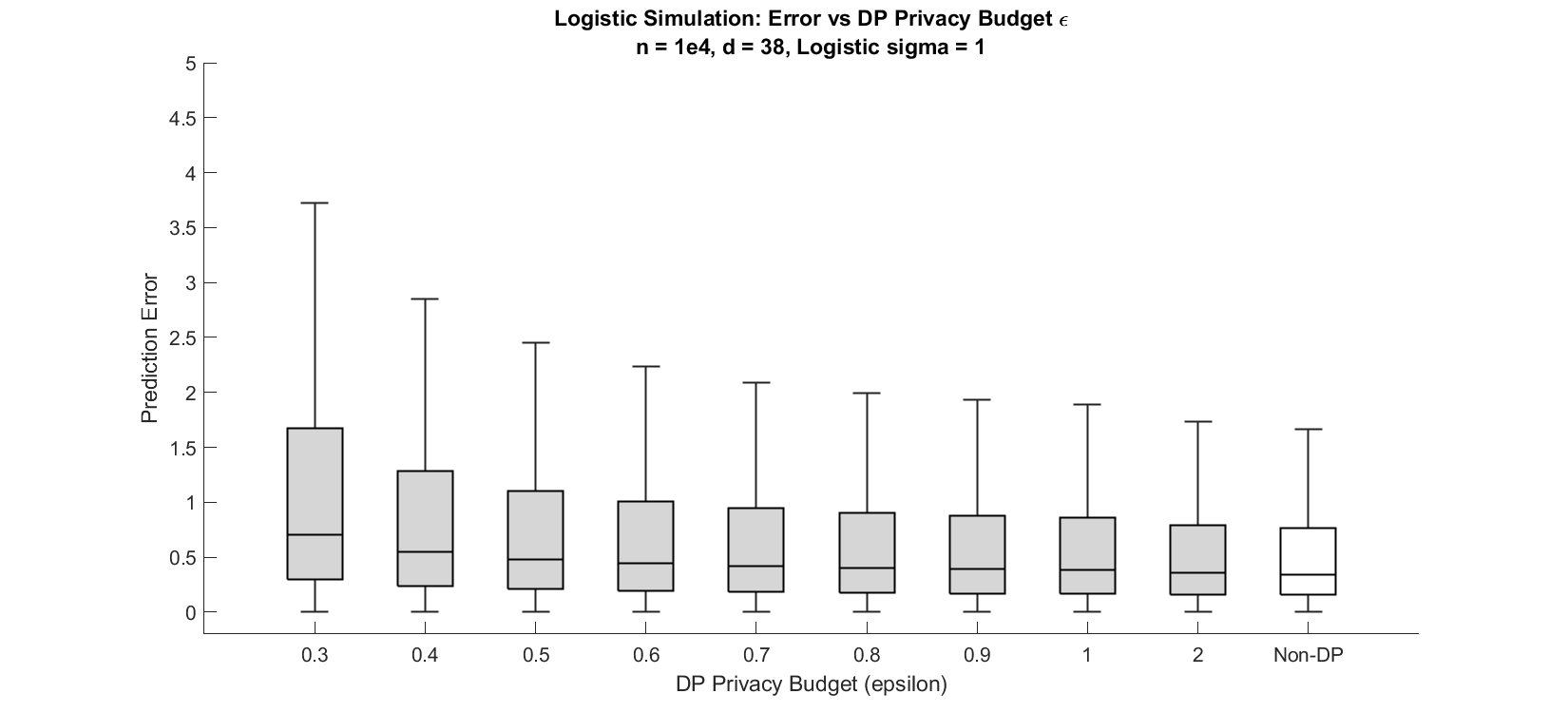}
\caption{Logistic Regression: Error vs DP privacy budget $\epsilon$}
\end{figure}

Figures 5 and 6 display the performance of DP models with varying privacy budgets $\epsilon$, with the rightmost boxplot representing the non-DP model. These figures illustrate that as the value of $\epsilon$ decreases (i.e., a stronger privacy guarantee is achieved), a higher prediction error is observed. Conversely, when the privacy budget becomes larger, the performance of the DP models is improved. For instance, when \(\epsilon = 0.3\), the median error for the differentially private SEV and Logistic regression models is approximately \(0.72\) and \(0.70\), respectively. When $\epsilon$ increases to $1$, the median errors of the two models both decrease to \(0.38\). This is reasonable since the privacy budget affects the magnitude of the noise introduced to the LLS regression models. Recall that the noises added into the DP model are determined by \(\operatorname{Lap}\left((4 + 4\sqrt{d} + d)/\epsilon)\right)\) for the SEV regression and \(\operatorname{Lap}\left((2 + 2\sqrt{d} + \frac{1}{2}d)/\epsilon\right)\) for Logistic regression. Therefore, when the value of \(d\) is fixed, the smaller \(\epsilon\) is, the larger the noises added, and consequently, the worse the model performs.

Figures 5 and 6 also demonstrate that the performance of the proposed differentially private LLS regression models converges with the increase in privacy budget. Both figures show that the prediction errors decrease rapidly when $\epsilon$ changes from smaller values to $1$. However, when $\epsilon$ is larger than $1$, the prediction errors do not exhibit a noticeable change and are almost identical to the non-DP models. We believe this is because the noise added to the regression model is sufficiently small (in comparison to the given sample size and feature dimension) when $\epsilon\geq 1$. As a result, the performance of the proposed models is not further compromised.

\section{Case Study} \label{sec:case}

In the case study, we utilize a dataset sourced from \cite{Saxena2008} to evaluate the performance of the proposed differentially private LLS regression. 

\subsection{Dataset and Experimental Settings}

The dataset comprises multi-sensor degradation data collected from aircraft turbofan engines. It consists of degradation signals from 100 training engines that ran to failure, degradation signals from an additional 100 test engines with operations prematurely terminated at random time points before failure, and the actual times-to-failure (TTF) corresponding to the 100 test engines. Each engine was monitored by 21 sensors. As an illustration, the degradation signals from one of the engines in the training dataset are shown in Figure 7. 
\begin{figure}[h]
\centering
\includegraphics[scale=0.30]{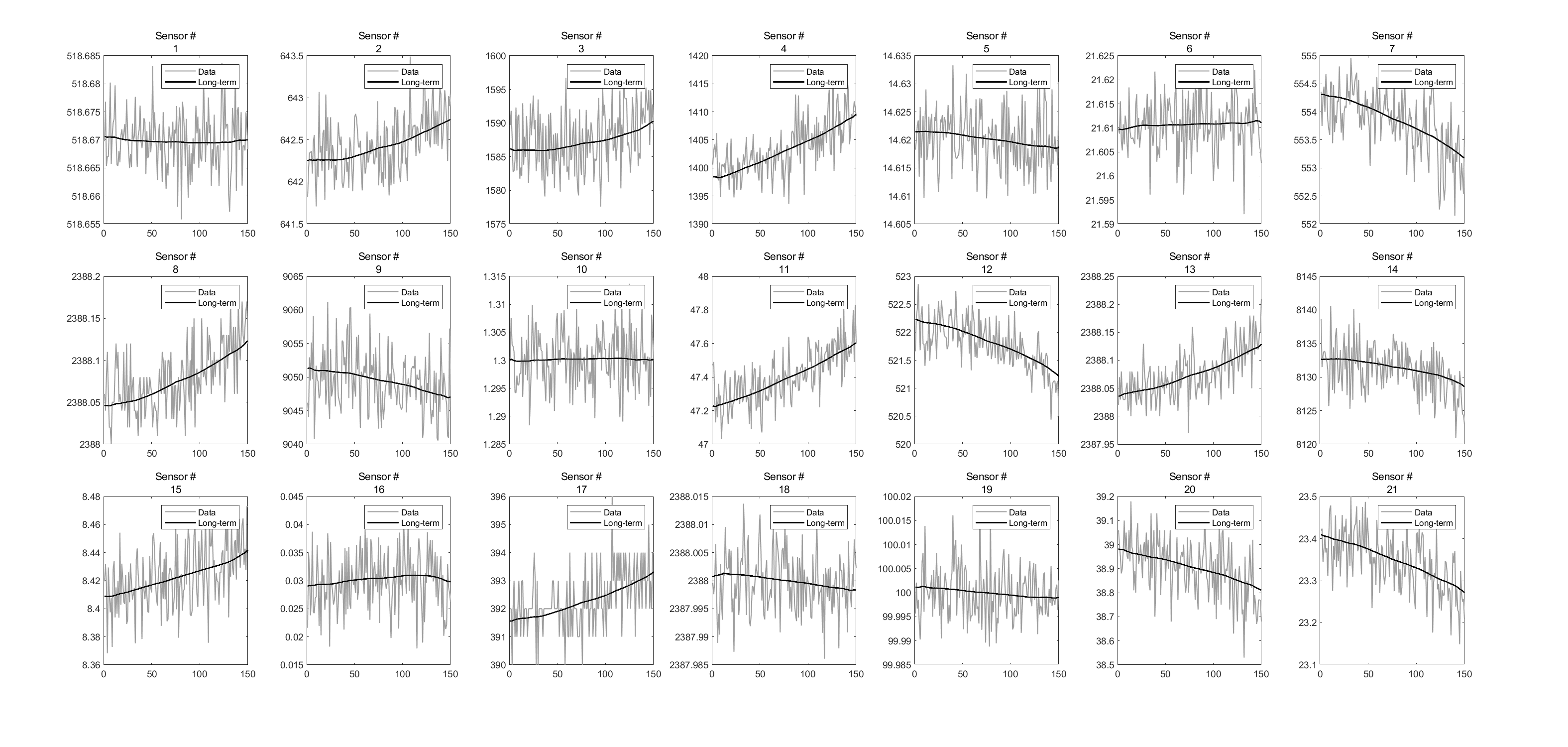}
\caption{Illustrated degradation signals from one of the engines.}
\end{figure}
\cite{fang2017multistream} have demonstrated that sensors 4, 17, and 20, along with SEV regression, are the most effective combination in modeling this dataset. Therefore, this study uses these three sensors and evaluates the performance of the proposed differentially private LLS regression under SEV distribution. It is known that the degradation signals in the training and test datasets have varying lengths due to failure truncation. Specifically, as machines are stopped for maintenance or replacement upon failure, the degradation signals can only be observed up to the time point of failure. Additionally, since different machines have different failure times, the length of degradation signals varies from one machine to another. To address this varying-length challenge, \cite{fang2017multistream} suggested the use of a time-varying framework, which truncates all degradation signals to the same length. In this study, we truncate the degradation signals from all engines by retaining their first 150 timestamps. Any engines with TTF smaller than 150 are excluded. Consequently, there are 94 engines left in the training dataset and 37 engines left in the test dataset.

Since the degradation signals from the three sensors have considerable correlations, we first apply Principal Component Analysis (PCA) to fuse them following the suggestion of \cite{fang2017multistream}. Next, we regress the TTFs against the features extracted from PCA using LLS regression. Similar to the simulation study, we use a non-DP SEV regression as the benchmark. We will investigate the performance of the proposed differentially private SEV regression under different predictor dimensions (i.e., the number of principal components) and also different privacy budgets. Since the DP noise was drawn from the Laplace distribution randomly each time, in order to evaluate the robustness of the DP models against the randomness of noise addition, the whole evaluation process was repeated 500 times for DP models, and the aggregated errors from all the repetitions are reported.

\subsection{Results and Analysis}

\subsubsection{Prediction Error vs. Dimensionality}

The first experiment focuses on investigating the performance of both DP and non-DP models with varying predictor dimensions. Specifically, we fix the privacy budget \(\epsilon\) at \(5\), and try the first \(3, 4, 5,\) and \(6\) principle components. Figure 8 shows the performance of both the non-DP model and the DP models.

\begin{figure}[h]
\centering
\includegraphics[scale=0.5]{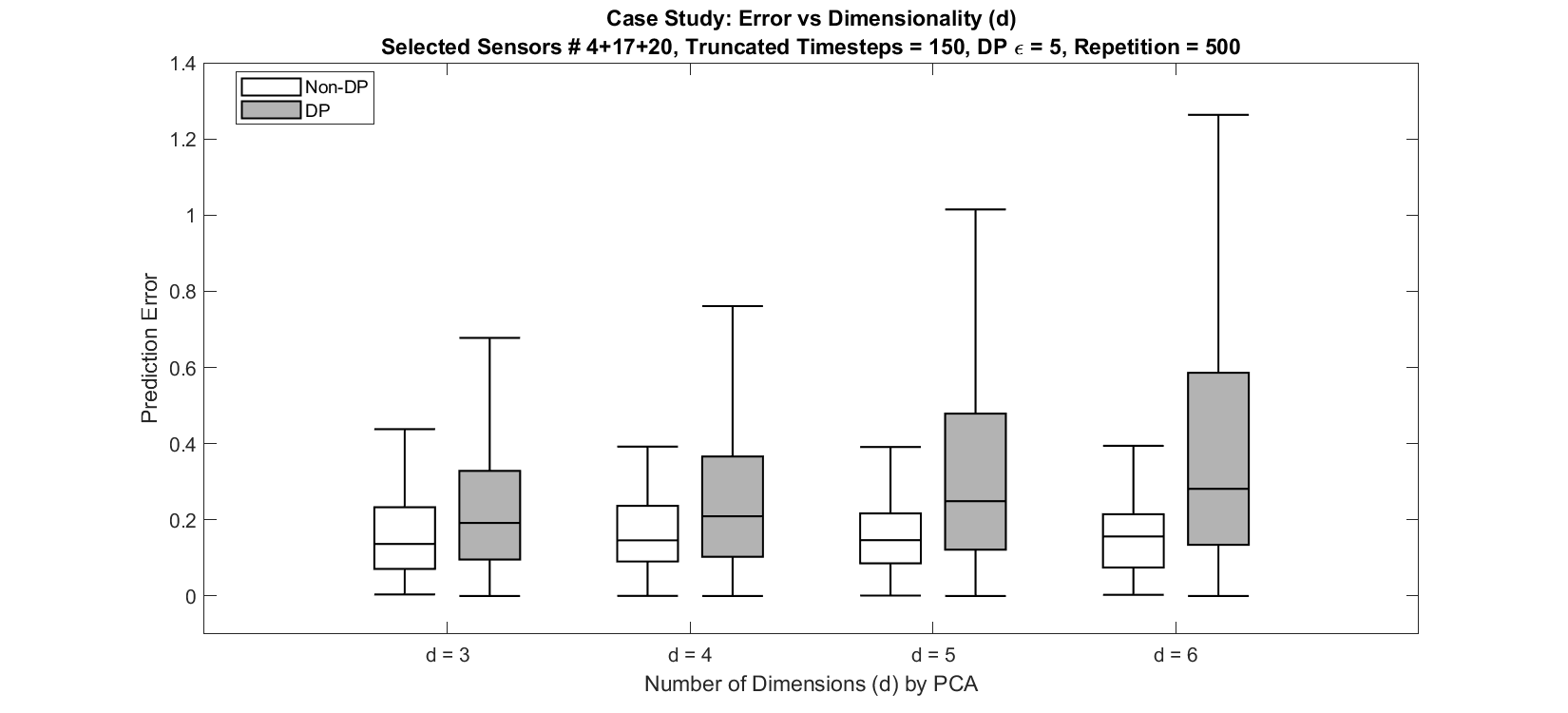}
\caption{Error vs Dimensionality}
\end{figure}

Figure 8 illustrates that the prediction errors of the DP model increase with the expansion of predictor dimensions, while the performance of the non-DP model remains stable across the various predictor dimensions tested. For example, the median error (and Interquartile Range, IQR) of the DP model at \(d=3\) is around \(0.19(0.24)\). However, it reaches \(0.21(0.0.26)\) at \(d=4\) and \(0.28(0.45)\) at \(d=6\). In contrast, the prediction errors (and IQR) of the non-DP model when $d=3,4,5$ and $6$ do not have a significant change and remain around $0.15(0.15)$.

We attribute the performance degradation of the DP model with the increase in predictor dimension to the heightened magnitude of noise injected into the model. As the number of features increases, the error correspondingly grows, which compromises the model's performance. Recall that the noise added to the differentially private SEV model is from \(\operatorname{Lap}\left((4 + 4\sqrt{d} + d)/\epsilon\right)\), which is a function of $d$. Thus, the noise added to the model increases with the increase of $d$. Another factor contributing to the inferior performance of the DP model compared to the non-DP model is the limited sample size in the training dataset. The simulation studies in Section \ref{sec:sim} have highlighted the significance of having sufficiently large training samples for DP models. In this study, the training sample size is 97, which is relatively small.

\subsubsection{Prediction Error vs. Privacy Budget}

In this subsection, we explore the influence of the privacy budget $\epsilon$ on the performance of the DP model. Here, we set the dimension number at $d=3$ and test a broad range of $\epsilon$ values, ranging from \(0.3\) to \(1\) that increment by $0.1$, and $1$, $1.5$, $2$, $3$, $4$, $5$, and $10$. Our objective is to observe any trends in the prediction errors of the DP model and examine whether the errors of the DP model converge with those of the non-DP model as $\epsilon$ increases.

\begin{figure}[H]
\centering
\includegraphics[scale=0.5]{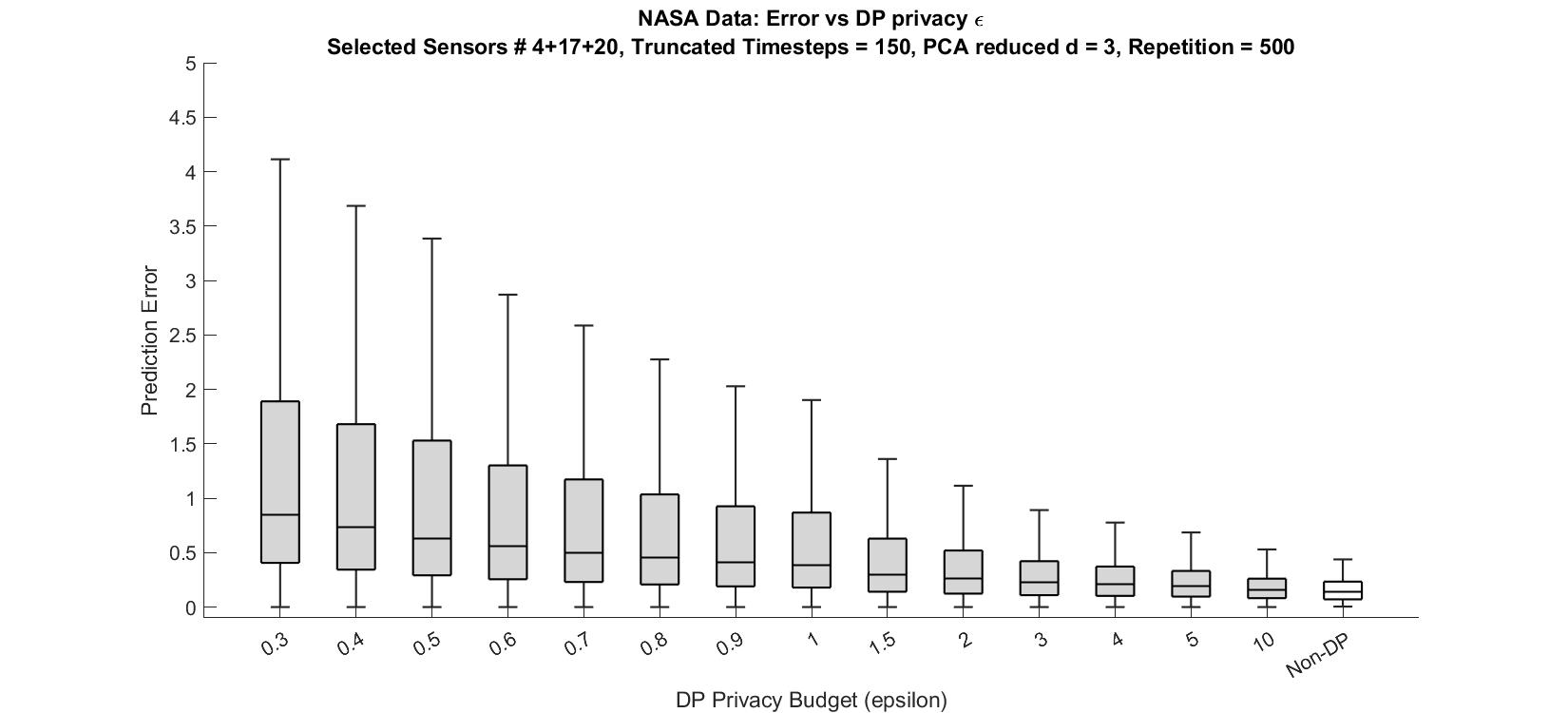}
\caption{Error vs DP budget $\epsilon$}
\end{figure}

Figure 9 depicts the results of experiments on the DP budget $\epsilon$, where the rightmost boxplot represents the error results associated with the non-DP model. It is evident that the errors of DP models consistently and clearly converge towards the non-DP errors as the value of $\epsilon$ increases. The median error values of the DP model are \(0.63\) at \(\epsilon=0.5\), decreasing to \(0.45\) at \(\epsilon=0.8\), and further to  \(0.38\) at \(\epsilon=1\). It keeps decreasing to equal or less than \(0.2\) when \(\epsilon \geq 5\), which is pretty close to the non-DP model. This observation suggests that the DP model with a larger $\epsilon$ value is more closely aligned with the performance of the non-DP model, while larger $\epsilon$ implies weaker privacy protection. This is reasonable since larger $\epsilon$ means less noise introduced to the differentially private LLS regression model.

\section{Conclusions}\label{sec:conclusions}
This article proposed differentially private (log)-location-scale regression. We incorporated differential privacy into two specific LLS regression models, SEV and Logistics, which can also be applied to Weibull regression and Log-logistic regression. The integration of differential privacy into the aforementioned models is based on the function mechanism, which works as follows. First, we employ Taylor expansion to decompose the log-likelihood function of LLS regression as a weighted combination of polynomial functions, truncating it at the second order. Next, random noise from the Laplace distribution is injected into the weights. Third, a perturbed log-likelihood function is reconstructed using the noisy weights and the polynomial functions. Finally, the perturbed log-likelihood function is solved for parameter estimation. We derived the sensitivities of the proposed DP models, which determine the magnitude of noise injected into the weights. Also, we proved that the proposed DP-LLS regression models satisfy $\epsilon$-differential privacy.

Extensive simulations and case studies were carried out to assess the performance of the proposed DP-LLS regression models. We focused on understanding how the size of the training dataset, sensor correlation, and privacy budget influence the DP method's effectiveness. The findings highlighted the significant impact of all three factors on the performance of the DP-LLS. Specifically, the results indicate that performance degrades with an increase in predictor dimension when keeping the sample size and privacy budget fixed. Performance improves with an increase in training sample size and stabilizes if the sample size is large enough while keeping the other two factors fixed. Furthermore, performance worsens with a decrease in the privacy budget (indicating stronger privacy protection) when keeping the predictor dimension and sample size fixed. In general, a sufficiently large training dataset is needed to ensure satisfactory prediction performance of the DP-LLS regression models and a desired level of privacy protection.

\section*{Appendix}\label{sec:appendix}

\subsection*{Proof of Proposition 1}

Let $\displaystyle \log q =\sum_{\substack{r=1}}^{\infty}\left[\frac{(-1)^{r+1}}{r}(q-1)^{r}\right]$ and expand it around $1$ in Taylor expansion, and let $\displaystyle e^x =\sum_{r=0}^{\infty} \frac{x^r}{r !}$ and expand it around $0$. We have
$$
\begin{aligned}
\tilde{\ell}\left(\{p_j\}_{j=0}^d,q\right) &=n \log q+\sum_{i=1}^n\left(y_i q-\sum_{j=0}^d p_j x_{i j}\right)-\sum_{i=1}^n \exp \left(y_i q-\sum_{j=0}^{d} p_j x_{i j}\right)\\
&=n \sum_{r=1}^{\infty} \frac{(-1)^{r+1}}{r}(q-1)^r+\sum_{i=1}^n y_i q-\sum_{i=1}^n \sum_{j=0}^d x_{i j} p_j -\sum_{i=1}^n \sum_{r=0}^{\infty} \frac{\left(y_i q-\sum_{j=0}^d p_j x_{i j}\right)^r}{r !}
\end{aligned}
$$
which is a polynomial of $p_j$ and $q$.
To truncate it to the second order, we have 
$$
\begin{aligned}
\tilde{\ell}\left(\{p_j\}_{j=0}^d,q\right) = & \quad n \sum_{r=1}^{2} \frac{(-1)^{r+1}}{r}(q-1)^r+\sum_{i=1}^n y_i q-\sum_{i=1}^n \sum_{j=0}^d x_{i j} p_j -\sum_{i=1}^n \sum_{r=0}^{2} \frac{\left(y_i q-\sum_{j=0}^d p_j x_{i j}\right)^r}{r !}\\
= & \quad n \left(\frac{(-1)^2}{1} (q-1)^1 + \frac{(-1)^3}{2}(q-1)^2\right) +\sum_{i=1}^n y_i q - \sum_{i=1}^n \sum_{j=0}^d x_{i j} p_j \\
&\quad - \sum_{i=1}^n \left(1+\frac{\left(y_i q-\sum_{j=0}^d p_j x_{i j}\right)^1}{1 !} + \frac{\left(y_i q-\sum_{j=0}^d p_j x_{i j}\right)^2}{2 !}\right) \\
=& \quad n\left((q-1)-\frac{1}{2}(q-1)^2\right) + \sum_{i=1}^n y_i q - \sum_{i=1}^n \sum_{j=0}^d x_{i j} p_j - n - \sum_{i=1}^n \left(y_i q-\sum_{j=0}^d p_j x_{i j}\right) \\
& \quad - \frac{1}{2} \sum_{i=1}^n \left(y_i q-\sum_{j=0}^d p_j x_{i j}\right)^2\\
\end{aligned}
$$
$$
\begin{aligned}
= & \quad  - \frac{3n}{2} + 2nq - \frac{n}{2} q^2 + \sum_{i=1}^n y_i q - \sum_{i=1}^n \sum_{j=0}^d x_{i j} p_j - n - \sum_{i=1}^n y_i q+\sum_{i=1}^n\sum_{j=0}^d p_j x_{i j} \\
&\quad - \frac{1}{2} \sum_{i=1}^n \left(y_i^2 q^2 - 2 y_i q \sum_{j=0}^d p_j x_{i j} + \left(\sum_{j=0}^d p_j x_{i j}\right)^2 \right)\\
= & \quad  - \frac{5n}{2} + 2nq - \frac{n}{2} q^2 - \frac{1}{2} \sum_{i=1}^n y_i^2 q^2 + \sum_{i=1}^n\left(y_i q \sum_{j=0}^d p_j x_{i j}\right) - \frac{1}{2} \sum_{i=1}^n \left(\sum_{j=0}^d p_j x_{i j}\right)^2 \\
= & \quad  - \frac{5n}{2} + 2nq - \frac{n}{2} q^2 - \frac{1}{2} \sum_{i=1}^n y_i^2 q^2 + \sum_{i=1}^n\sum_{j=0}^d y_i x_{i j} p_j q  - \frac{1}{2} \sum_{i=1}^n \left( \sum_{j=0}^d p_j^2 x_{i j}^2 + \left(\sum_{h=0, h\neq j}^d \sum_{j=0}^d p_j p_h x_{i j} x_{i h}  \right)\right)\\
= & \quad  - \frac{5n}{2} + 2nq - \frac{n}{2} q^2  - \frac{1}{2} \sum_{i=1}^n y_i^2 q^2 + \sum_{i=1}^n\sum_{j=0}^d y_i x_{i j} p_j q - \frac{1}{2} \sum_{i=1}^n \sum_{j=0}^d x_{i j}^2 p_j^2 -\frac{1}{2} \sum_{i=1}^n \left(\sum_{h=0, h\neq j}^d \sum_{j=0}^d x_{i j} x_{i h} p_j p_h \right)\\
\end{aligned}
$$

\subsection*{Proof of Proposition 2}
Recall the polynomial function
\begin{multline*}
 \tilde{\ell}\left(\{p_j\}_{j=0}^d,q\right) =    - \frac{5n}{2} + 2nq- \frac{1}{2}\left(n+ \sum_{i=1}^n y_i^2\right) q^2   + \sum_{i=1}^n\sum_{j=0}^d y_i x_{i j} p_j q - \frac{1}{2} \sum_{i=1}^n \sum_{j=0}^d x_{i j}^2 p_j^2 \\-\frac{1}{2}\sum_{i=1}^n \left(\sum_{h=0, h\neq j}^d \sum_{j=0}^d x_{i j} x_{i h} p_j p_h \right).   
\end{multline*}

Notice we have $d+2$ variables, $\{p_0, p_1, \dots, p_d, q \}$. Since our polynomial function was truncated by the second order, the polynomial basis of this function consists of zeroth-order, first-order, and second-order combinations of the set of variables. The zeroth-order basis is ${1}$ and its corresponding coefficient is $w_1 = - \frac{5n}{2} $. The first-order basis is $q$ and its corresponding coefficient is $w_q = 2n$. The second-order basis contains $q^2$, $\{p_j q\}_{j=0}^d$, $\{p_j ^2\}_{j=0}^d$, and the pairwise terms $\{\{p_j p_h\}_{j=0}^d\}_{h=0,h\neq j}^d$. Their corresponding coefficients are $w_{q^2} = - \frac{1}{2}\left(n+ \sum_{i=1}^n y_i^2\right)$, $\{w_{p_j q}=\sum_{i=1}^n y_i x_{i j}\}_{j=0}^d,\{w_{p_j^2}=- \frac{1}{2} \sum_{i=1}^n x_{i j}^2\}_{j=0}^d$, and $\{\{w_{p_j p_h}=-\frac{1}{2}\sum_{i=1}^n x_{i j} x_{i h}\}_{j=0}^d\}_{h=0,h\neq j}^d$, respectively. 

Let $D$ and $D^{\prime}$ be any two neighbor databases that differ by one data point. Without loss of generality, we assume that $D$ and $D'$ differ in the last sample. Let $(y_n, \{x_{nj}\}_{j=1}^d)$ be the last sample in $D$ and $(y_n', \{x_{nj}'\}_{j=1}^d)$ be the last sample in $D'$. Recall that $\tilde{\ell}_D\left(\left\{p_j\right\}_{j=0}^d, q\right)$ and $\tilde{\ell}_{D^{\prime}}\left(\left\{p_j\right\}_{j=0}^d, q\right)$ are the objective functions of regression analysis on $D$ and $D^{\prime}$, respectively. In addition, the weights of $\tilde{\ell}_D\left(\left\{p_j\right\}_{j=0}^d, q\right)$ are $w_1^D$, $w_q^D,w_{q^2}^D$, $\{w_{p_jq}^D\}_{j=1}^d$,$\{w_{p_j^2}^D\}_{j=0}^d$, $\{\{w_{p_jp_h}^D\}_{j=0}^d\}_{h=0,h\neq j}^d$ and the weights of $\tilde{\ell}_{D'}\left(\left\{p_j\right\}, q\right)$ are $w_1^{D'}$, $w_q^{D'},w_{q^2}^{D'}$,$\{w_{p_jq}^{D'}\}_{j=1}^d$,$\{w_{p_j^2}^{D'}\}_{j=0}^d$, $\{\{w_{p_jp_h}^{D'}\}_{j=0}^d\}_{h=0,h\neq j}^d$. Thus, we have 
$$
\begin{aligned}
&\| w_1^D - w_1^{D^{\prime}} \|_1  + \| w_q^D - w_q^{D^{\prime}} \|_1 + \| w_{q^2}^D - w_{q^2}^{D^{\prime}} \|_1 + \sum_{j=0}^d \| w_{p_j q}^D - w_{p_j q}^{D^{\prime}} \|_1  + \sum_{j=0}^d \| w_{p_j^2}^D - w_{p_j^2}^{D^{\prime}} \|_1 + \\&\quad\quad\quad\quad\quad\sum_{n=1,h \neq j}^d \sum_{j=0}^d \| w_{p_j p_h}^D - w_{p_j p_h}^{D^{\prime}} \|_1\\
&= 0+0+\|-\frac{1}{2}y_n^2+\frac{1}{2}y_{n}'^2\|_1+\sum_{j=0}^d\|y_nx_{nj}-y_n'x_{nj}'\|_1+\sum_{j=0}^d\|-\frac{1}{2}x_{nj}^2+\frac{1}{2}x_{nj}'^2\|_1+\\&\quad\quad\quad\quad\quad\sum_{j=0}^d\sum_{h=0, h \neq j}^d \frac{1}{2}\|-x_{nj}x_{nh}+x_{nj}'x_{nh}'\|_1\\
&\leq \|-\frac{1}{2}y_n^2\|+\|\frac{1}{2}y_{n}'^2\|_1+\sum_{j=0}^d\|y_nx_{nj}\|_1+\sum_{j=0}^d\|-y_n'x_{nj}'\|_1+\sum_{j=0}^d\|-\frac{1}{2}x_{nj}^2\|_1+\sum_{j=0}^d\|\frac{1}{2}x_{nj}'^2\|_1\\&\quad\quad\quad\quad\quad\sum_{j=0}^d\sum_{h=0, h \neq j}^d \frac{1}{2}\|-x_{nj}x_{nh}\|_1+\sum_{j=0}^d\sum_{h=0, h \neq j}^d \frac{1}{2}\|x_{nj}'x_{nh}'\|_1\\
&\leq \frac{1}{2}+\frac{1}{2}+(1+\frac{1}{\sqrt{d}}\times d)+(1+\frac{1}{\sqrt{d}}\times d)+\frac{1}{2}(1+1)+\frac{1}{2}(1+1)+\\&\quad\quad\quad\quad\quad+\frac{1}{2}(d\times (d-1)\times \frac{1}{d}+2\times d\times \frac{1}{\sqrt{d}})+\frac{1}{2}(d\times (d-1)\times \frac{1}{d}+2\times d\times \frac{1}{\sqrt{d}})\\
&=4+4\sqrt{d}+d
\end{aligned}
$$

Please notice that we have scaled the $j$th predictor by using $x_{i j}=\frac{\tilde{x}_{i j}-\alpha_j}{\left(\beta_j-\alpha_j\right) \cdot \sqrt{d}}$, where $\alpha_j=\min\{\{\tilde{x}_{ij}\}_{i=1}^n\}$ and $\beta_j=\max\{\{\tilde{x}_{ij}\}_{i=1}^n\}$ denotes the minimum and maximum values of the $j$th predictor, respectively. Thus, we have $\|x_{nj}\|_1\leq \frac{1}{\sqrt{d}}$, $\|x_{nj}'\|_1\leq \frac{1}{\sqrt{d}}$, $\|x_{nh}\|_1\leq \frac{1}{\sqrt{d}}$, $\|x_{nh}'\|_1\leq \frac{1}{\sqrt{d}}$, $\|x_{nj}^2\|_1\leq \frac{1}{d}$, $\|x_{nj}'^2\|_1\leq \frac{1}{d}$, $\sum_{i=1}^d\|x_{nj}^2\|\leq 1$, $\|x_{nj}x_{nh}\|_1\leq \frac{1}{d}$. Also, recall that we have scaled $y_i \in [-1,1]$ for all $i$. Thus, $\|y_n\|_1\leq 1$, $\|y_n'\|\leq 1$, $\|y_nx_{nj}\|_1\leq\frac{1}{\sqrt{d}}$, and $\|y_n'x_{nj}'\|_1\leq\frac{1}{\sqrt{d}}$. Furthermore, we have not scaled the $x_{n0}$, which is the predictor corresponding to the intecept. Thus, $\|x_{n0}\|_1=1$.

\subsection*{Proof of Proposition 3}
Let $D$ and $D^{\prime}$ be any two neighbor databases that differ by one data point. Without loss of generality, we assume that $D$ and $D'$ differ in the last sample. Let $(y_n, \{x_{nj}\}_{j=1}^d)$ be the last sample in $D$ and $(y_n', \{x_{nj}'\}_{j=1}^d)$ be the last sample in $D'$.  Let $\mathcal{P}(\tilde{\tilde{\ell}}\left(\left\{p_j\right\}_{j=0}^d, q\right)|D)$ and $\mathcal{P}(\tilde{\tilde{\ell}}\left(\left\{p_j\right\}_{j=0}^d, q\right)
|D^{\prime})$ denote the probability that the output of the algorithm given dataset D and $D^{\prime}$ respectively. Let $\tilde{w}$ denote polynomial weights after adding noises, i.e.,  $w+ \operatorname{Lap}(\frac{\Delta}{\epsilon})$. For example, $\tilde{w}_1=w_1^D+ \operatorname{Lap}(\frac{\Delta}{\epsilon})$ using database $D$ and $\tilde{w}_1=w_1^{D'}+ \operatorname{Lap}(\frac{\Delta}{\epsilon})$ using database $D'$. Recall that the pdf of Laplace distribution with zero mean and scale $\frac{\Delta}{\epsilon}$ is $pdf(x)=\frac{\epsilon}{2\Delta}\exp(-\frac{\epsilon\|x\|_1}{\Delta})$ and $\Delta=4+4\sqrt{d}+d$. Thus, we have the following:

$$
\begin{aligned}
&\frac{\mathcal{P}(\tilde{\tilde{\ell}}\left(\left\{p_j\right\}_{j=0}^d, q\right)|D)}{\mathcal{P}(\tilde{\tilde{\ell}}\left(\left\{p_j\right\}_{j=0}^d, q\right)
|D^{\prime})} \\
&=\frac{\frac{\epsilon}{2\Delta}\exp\left(-\frac{\epsilon\|\tilde{w}_1-w_1^{D}\|_1}{\Delta}\right)}{\frac{\epsilon}{2\Delta}\exp\left(-\frac{\epsilon\|\tilde{w}_1-w_1^{D'}\|_1}{\Delta}\right)}\frac{\frac{\epsilon}{2\Delta}\exp\left(-\frac{\epsilon\|\tilde{w}_q-w_q^{D}\|_1}{\Delta}\right)}{\frac{\epsilon}{2\Delta}\exp\left(-\frac{\epsilon\|\tilde{w}_q-w_q^{D'}\|_1}{\Delta}\right)}\frac{\frac{\epsilon}{2\Delta}\exp\left(-\frac{\epsilon\|\tilde{w}_{q^2}-w_{q^2}^{D}\|_1}{\Delta}\right)}{\frac{\epsilon}{2\Delta}\exp\left(-\frac{\epsilon\|\tilde{w}_{q^2}-w_{q^2}^{D'}\|_1}{\Delta}\right)}\\
& \quad\quad\quad\quad\quad\frac{\prod_{j=0}^d\frac{\epsilon}{2\Delta}\exp\left(-\frac{\epsilon\|\tilde{w}_{p_jq}-w_{p_jq}^{D}\|_1}{\Delta}\right)}{\prod_{j=0}^d\frac{\epsilon}{2\Delta}\exp\left(-\frac{\epsilon\|\tilde{w}_{p_jq}-w_{p_jq}^{D'}\|_1}{\Delta}\right)}\frac{\prod_{j=0}^d\frac{\epsilon}{2\Delta}\exp\left(-\frac{\epsilon\|\tilde{w}_{p_j^2}-w_{p_j^2}^{D}\|_1}{\Delta}\right)}{\prod_{j=0}^d\frac{\epsilon}{2\Delta}\exp\left(-\frac{\epsilon\|\tilde{w}_{p_j^2}-w_{p_j^2}^{D'}\|_1}{\Delta}\right)}\\
& \quad\quad\quad\quad\quad\frac{\prod_{j=0}^d\prod_{h=0,h\neq j}^d\frac{\epsilon}{2\Delta}\exp\left(-\frac{\epsilon\|\tilde{w}_{p_jp_h}-w_{p_jp_h}^{D}\|_1}{\Delta}\right)}{\prod_{j=0}^d\prod_{h=0,h\neq j}^d\frac{\epsilon}{2\Delta}\exp\left(-\frac{\epsilon\|\tilde{w}_{p_jp_h}-w_{p_jp_h}^{D'}\|_1}{\Delta}\right)}\\
& \leq \exp \left(\frac{\epsilon}{\Delta}(\| \tilde{w}_1^D - \tilde{w}_1^{D^{\prime}} \|_1  + \| \tilde{w}_q^D - \tilde{w}_q^{D^{\prime}} \|_1 + \| \tilde{w}_{q^2}^D - \tilde{w}_{q^2}^{D^{\prime}} \|_1 +\sum_{j=0}^d \| \tilde{w}_{p_j q}^D - \tilde{w}_{p_j q}^{D^{\prime}} \|_1 + \right.\\
&  \quad\quad\quad\quad\quad\left.\sum_{j=0}^d \| \tilde{w}_{p_j^2}^D - \tilde{w}_{p_j^2}^{D^{\prime}} \|_1 + \sum_{j=0 }^d \sum_{h=0,h\neq j}^d \| \tilde{w}_{p_j p_h}^D - \tilde{w}_{p_j p_h}^{D^{\prime}} \|_1 )\right)\\
& \leq  \exp \left(\frac{\epsilon}{\Delta}\left(4+4\sqrt{d}+d\right)\right)\\
& = \exp(\epsilon)
\end{aligned}
$$

This implies that the computation of $\tilde{\tilde{\ell}}\left(\left\{p_j\right\}_{j=0}^d, q\right)$ ensures $\epsilon$-differential privacy. Since the final result of Algorithm \ref{alg:sev} is achieved by solving $\tilde{\tilde{\ell}}\left(\left\{p_j\right\}_{j=0}^d, q\right)$ without using any additional information from the original database, Algorithm \ref{alg:sev} satisfies $\epsilon$-differential privacy as well. 

\subsection*{Proof of Proposition 4}
Based on Taylor's expansion, we have 
$$
\log \left(1+e^z\right)\approx\left[\log (2)+\frac{z}{2}+\frac{z^2}{8}\right]
$$
Recall that  $l(\tilde{\mu}, \sigma)$ can be written using as follows:
$$
\begin{aligned}
l\left(\{p_j\}_{j=0}^d,q\right) &=n \log q+\sum_{i=1}^n\left(y_i q-\sum_{j=0}^d p_j x_{i j}\right)-2\sum_{i=1}^n \log\left(1+ \exp \left(y_i q-\sum_{j=0}^{d} p_j x_{i j}\right)\right)
\end{aligned}
$$
To truncate it to the second order, we have:

$$
\begin{aligned}
&\tilde{\ell}\left(\{p_j\}_{j=0}^d,q\right) \\
= & \quad  n \sum_{r=1}^{2} \frac{(-1)^{r+1}}{r}(q-1)^r+\sum_{i=1}^n y_i q-\sum_{i=1}^n \sum_{j=0}^d x_{i j} p_j \\
&\quad -2\sum_{i=1}^n \left[\log2 + \frac{1}{2} (y_i q-\sum_{j=0}^d p_j x_{i j}) + \frac{1}{8} (y_i q-\sum_{j=0}^d p_j x_{i j})^2\right]\\
= & \quad   - \frac{3n}{2} + 2nq - \frac{n}{2} q^2+ \sum_{i=1}^n y_i q - \sum_{i=1}^n \sum_{j=0}^d x_{i j} p_j - 2n\log 2 \\
&\quad - \sum_{i=1}^n y_i q + \sum_{i=1}^n \sum_{j=0}^d p_j x_{i j} - \frac{1}{4} \sum_{i=1}^n \left(y_i q - \sum_{j=0}^d p_j x_{i j}\right)^2\\\
= & \quad  - \frac{3n}{2} + 2nq - \frac{n}{2} q^2- 2n\log 2 - \frac{1}{4} \sum_{i=1}^n \left(y_i^2 q^2 - 2 y_i q \sum_{j=0}^d p_j x_{i j} + \left(\sum_{j=0}^d p_j x_{i j}\right)^2 \right)\\
= & \quad  - \frac{3n}{2} + 2nq - \frac{n}{2} q^2- 2n\log 2-\frac{1}{4} \sum_{i=1}^n y_i^2 q^2 + \frac{1}{2}\sum_{i=1}^n\left(y_i q \sum_{j=0}^d p_j x_{i j}\right) - \frac{1}{4} \sum_{i=1}^n \left(\sum_{j=0}^d p_j x_{i j}\right)^2\\
= & \quad  - \frac{3n}{2} + 2nq - \frac{n}{2} q^2- 2n\log 2 - \frac{1}{4} \sum_{i=1}^n y_i^2 q^2 + \frac{1}{2}\sum_{i=1}^n\sum_{j=0}^d y_i x_{i j} p_j q  \\
&\quad - \frac{1}{4} \sum_{i=1}^n \left( \sum_{j=0}^d p_j^2 x_{i j}^2 + \left(\sum_{h\neq j}^d \sum_{j=0}^d p_j p_h x_{i j} x_{i h}  \right)\right)\\
\end{aligned}
$$

$$
\begin{aligned}
= & \quad - \frac{3n}{2} + 2nq - \frac{n}{2} q^2-  2n\log 2 - \frac{1}{4} \sum_{i=1}^n y_i^2 q^2 + \frac{1}{2} \sum_{i=1}^n\sum_{j=0}^d y_i x_{i j} p_j q \\
&\quad - \frac{1}{4} \sum_{i=1}^n \sum_{j=0}^d x_{i j}^2 p_j^2 - \frac{1}{4}\sum_{i=1}^n \left(\sum_{h\neq j}^d \sum_{j=0}^d x_{i j} x_{i h} p_j p_h \right)\\
\end{aligned}
$$

\subsection*{Proof of Proposition 5}
Recall our polynomial function is 
$$
\begin{aligned}
\tilde{\ell}\left(\{p_j\}_{j=0}^d,q\right) = & \quad -n \left(\frac{3}{2}+ 2\log 2\right) + 2nq - \left(\frac{n}{2}+\frac{1}{4} \sum_{i=1}^n y_i^2 \right) q^2  + \frac{1}{2} \sum_{i=1}^n\sum_{j=0}^d y_i x_{i j} p_j q \\
&\quad - \frac{1}{4} \sum_{i=1}^n \sum_{j=0}^d x_{i j}^2 p_j^2 - \frac{1}{4}\sum_{i=1}^n \left(\sum_{h\neq j}^d \sum_{j=0}^d x_{i j} x_{i h} p_j p_h \right)\\
\end{aligned}
$$
The zeroth-order basis $1$ has coefficient $w_1 =  -n \left(\frac{3}{2}+ 2\log 2\right)$, and the first-order basis $q$ has coefficient $w_q = 2n$. The second-order basis are $q^2$, $\{p_j q\}_{j=0}^d$, $\{p_j ^2\}_{j=0}^d$, and the pairwise combinations $\{\{p_j p_h\}_{j=0}^d\}_{h=0,h\neq j}^d$. Their corresponding coefficients are  $w_{q^2} =-\left(\frac{n}{2}+\frac{1}{4} \sum_{i=1}^n y_i^2\right) $, $\{w_{p_j q}=\frac{1}{2} \sum_{i=1}^ny_i x_{i j}\}_{j=0}^d$, $\{w_{p_j^2}=-\frac{1}{4} \sum_{i=1}^n x_{i j}^2\}_{j=0}^d$, and $\{\{w_{p_j p_h}=- \frac{1}{4}\sum_{i=1}^n x_{i j} x_{i h}\}_{j=0}^d\}_{h=0,h\neq j}^d$.

Let $D$ and $D^{\prime}$ be any two neighbor databases that differ by one data point. Without loss of generality, we assume that $D$ and $D'$ differ in the last sample. Let $(y_n, \{x_{nj}\}_{j=1}^d)$ be the last sample in $D$ and $(y_n', \{x_{nj}'\}_{j=1}^d)$ be the last sample in $D'$. Recall that $\tilde{\ell}_D\left(\left\{p_j\right\}_{j=0}^d, q\right)$ and $\tilde{\ell}_{D^{\prime}}\left(\left\{p_j\right\}_{j=0}^d, q\right)$ are the objective functions of regression analysis on $D$ and $D^{\prime}$, respectively. In addition, the weights of $\tilde{\ell}_D\left(\left\{p_j\right\}_{j=0}^d, q\right)$ are $w_1^D$, $w_q^D,w_{q^2}^D$, $\{w_{p_jq}^D\}_{j=1}^d$,$\{w_{p_j^2}^D\}_{j=0}^d$, $\{\{w_{p_jp_h}^D\}_{j=0}^d\}_{h=0,h\neq j}^d$ and the weights of $\tilde{\ell}_{D'}\left(\left\{p_j\right\}, q\right)$ are $w_1^{D'}$, $w_q^{D'},w_{q^2}^{D'}$,$\{w_{p_jq}^{D'}\}_{j=1}^d$,$\{w_{p_j^2}^{D'}\}_{j=0}^d$, $\{\{w_{p_jp_h}^{D'}\}_{j=0}^d\}_{h=0,h\neq j}^d$. Thus, we have 
$$
\begin{aligned}
&\| w_1^D - w_1^{D^{\prime}} \|_1  + \| w_q^D - w_q^{D^{\prime}} \|_1 + \| w_{q^2}^D - w_{q^2}^{D^{\prime}} \|_1 + \sum_{j=0}^d \| w_{p_j q}^D - w_{p_j q}^{D^{\prime}} \|_1  + \sum_{j=0}^d \| w_{p_j^2}^D - w_{p_j^2}^{D^{\prime}} \|_1 + \\&\quad\quad\quad\quad\quad\sum_{n=1,h \neq j}^d \sum_{j=0}^d \| w_{p_j p_h}^D - w_{p_j p_h}^{D^{\prime}} \|_1\\
&= 0+0+\|-\frac{1}{4}y_n^2+\frac{1}{4}y_{n}'^2\|_1+\frac{1}{2}\sum_{j=0}^d\|y_nx_{nj}-y_n'x_{nj}'\|_1+\sum_{j=0}^d\|-\frac{1}{4}x_{nj}^2+\frac{1}{4}x_{nj}'^2\|_1+\\&\quad\quad\quad\quad\quad\frac{1}{4}\sum_{j=0}^d\sum_{h=0, h \neq j}^d \|-x_{nj}x_{nh}+x_{nj}'x_{nh}'\|_1\\
&\leq \|-\frac{1}{4}y_n^2\|+\|\frac{1}{4}y_{n}'^2\|_1+\frac{1}{2}\sum_{j=0}^d\|y_nx_{nj}\|_1+\frac{1}{2}\sum_{j=0}^d\|-y_n'x_{nj}'\|_1+\sum_{j=0}^d\|-\frac{1}{4}x_{nj}^2\|_1+\sum_{j=0}^d\|\frac{1}{4}x_{nj}'^2\|_1\\&\quad\quad\quad\quad\quad\frac{1}{4}\sum_{j=0}^d\sum_{h=0, h \neq j}^d \|-x_{nj}x_{nh}\|_1+\frac{1}{4}\sum_{j=0}^d\sum_{h=0, h \neq j}^d \|x_{nj}'x_{nh}'\|_1\\
\end{aligned}
$$
$$
\begin{aligned}
&\leq \frac{1}{4}+\frac{1}{4}+\frac{1}{2}(1+\frac{1}{\sqrt{d}}\times d)+\frac{1}{2}(1+\frac{1}{\sqrt{d}}\times d)+\frac{1}{4}(1+1)+\frac{1}{4}(1+1)+\\&\quad\quad\quad\quad\quad\frac{1}{4}(d\times (d-1)\times \frac{1}{d}+2\times d\times \frac{1}{\sqrt{d}})+\frac{1}{4}(d\times (d-1)\times \frac{1}{d}+2\times d\times \frac{1}{\sqrt{d}})\\
&=2+2\sqrt{d}+\frac{1}{2}d
\end{aligned}
$$

Please notice that we have scaled the $j$th predictor by using $x_{i j}=\frac{\tilde{x}_{i j}-\alpha_j}{\left(\beta_j-\alpha_j\right) \cdot \sqrt{d}}$, where $\alpha_j=\min\{\{\tilde{x}_{ij}\}_{i=1}^n\}$ and $\beta_j=\max\{\{\tilde{x}_{ij}\}_{i=1}^n\}$ denotes the minimum and maximum values of the $j$th predictor, respectively. Thus, we have $\|x_{nj}\|_1\leq \frac{1}{\sqrt{d}}$, $\|x_{nj}'\|_1\leq \frac{1}{\sqrt{d}}$, $\|x_{nh}\|_1\leq \frac{1}{\sqrt{d}}$, $\|x_{nh}'\|_1\leq \frac{1}{\sqrt{d}}$, $\|x_{nj}^2\|_1\leq \frac{1}{d}$, $\|x_{nj}'^2\|_1\leq \frac{1}{d}$, $\sum_{i=1}^d\|x_{nj}^2\|\leq 1$, $\|x_{nj}x_{nh}\|_1\leq \frac{1}{d}$. Also, recall that we have scaled $y_i \in [-1,1]$ for all $i$. Thus, $\|y_n\|_1\leq 1$, $\|y_n'\|\leq 1$, $\|y_nx_{nj}\|_1\leq\frac{1}{\sqrt{d}}$, and $\|y_n'x_{nj}'\|_1\leq\frac{1}{\sqrt{d}}$. Furthermore, we have not scaled the $x_{n0}$, which is the predictor corresponding to the intercept. Thus, $\|x_{n0}\|_1=1$.

\subsection*{Proof of Proposition 6}
The proof is the same to that of the proof of Proposition 3 except that $\Delta=2+2\sqrt{d}+\frac{1}{2}d$.

\if1\blind{
\section*{Acknowledgements}
The authors acknowledge the generous support from the funding agency of XYZ.	} \fi

\bibliographystyle{chicago}
\spacingset{1}
\bibliography{IISE-Trans}
	
\end{document}